\documentclass[12pt,a4paper]{article}
\usepackage{amsmath,amsfonts}
\usepackage{hyperref}
\usepackage[numbers]{natbib}
\usepackage{multirow}
\usepackage{booktabs}
\usepackage{cases}
\usepackage[margin=1in]{geometry}
\usepackage{authblk}
\usepackage{graphicx}

\title{Hybrid Two-Stage Reconstruction of Multiscale Subsurface Flow with Physics-informed Residual Connected Neural Operator}

\author[1]{Peiqi Li}
\author[1,*]{Jie Chen}

\affil[1]{\small School of Mathematics and Physics, Xi'an Jiaotong-Liverpool University \\ \texttt{LPQ\_0619@outlook.com (Peiqi Li),  Jie.Chen01@xjtlu.edu.cn (Jie Chen)}}
\affil[*]{Corresponding Author.}

\begin{document}

\maketitle

\begin{abstract}
The novel neural networks show great potential in solving partial differential equations. For single-phase flow problems in subsurface porous media with high-contrast coefficients, the key is to develop neural operators with accurate reconstruction capability and strict adherence to physical laws. In this study, we proposed a hybrid two-stage framework that uses multiscale basis functions and physics-guided deep learning to solve the Darcy flow problem in high-contrast fractured porous media. In the first stage, a data-driven model is used to reconstruct the multiscale basis function based on the permeability field to achieve effective dimensionality reduction while preserving the necessary multiscale features. In the second stage, the physics-informed neural network, together with Transformer-based global information extractor is used to reconstruct the pressure field by integrating the physical constraints derived from the Darcy equation, ensuring consistency with the physical laws of the real world. The model was evaluated on datasets with different combinations of permeability and basis functions and performed well in terms of reconstruction accuracy. Specifically, the framework achieves R2 values above 0.9 in terms of basis function fitting and pressure reconstruction, and the residual indicator is on the order of $1\times 10^{-4}$. These results validate the ability of the proposed framework to achieve accurate reconstruction while maintaining physical consistency.
\end{abstract}


\newpage

\section{Introduction}
After decades of development, traditional numerical solvers (finite difference methods (FDM), finite element methods (FEM), etc.) have demonstrated outstanding effectiveness and robustness in solving complex partial differential equations (PDEs)\cite{kippe2008comparison}. However, when dealing with problems that require high iterations or parameter inversion, these traditional methods do not have a clear advantage in terms of efficiency\cite{efendiev2007multiscale}. Taking the problem of fluid flow in porous media, especially in fractured porous media with high contrast and heterogeneity, as an example, the solution process requires consideration of complex fracture systems\cite{aarnes2006adaptive}. The fractures in the system vary significantly in size and direction, causing the problem to span multiple scales. In such multiscale systems, using large-scale models may reduce the accuracy of the model due to the neglect of small fractures, while small-scale computations significantly increase computational costs. Therefore, the introduction of multiscale techniques is essential, providing an effective solution for simulating fluid flow in heterogeneous porous media\cite{dehkordi2013multi}.

Multiscale techniques have widespread applications in solving fluid flow problems. Common multiscale methods include the Multiscale Finite Volume Method (MsFVM)\cite{jenny2003multi,hajibeygi2008iterative}, Multiscale Finite Element Method (MsFEM)\cite{hou1997multiscale,efendiev2009multiscale}, and Mortar Multiscale Method\cite{arbogast2007multiscale,arbogast2013multiscale}. These methods solve problems by applying precomputed multiscale basis functions on coarse grids. Jenny et al.\cite{jenny2003multi}. proposed a finite volume method based on a flux-continuous finite difference scheme, which simplifies the flow problem by solving small-scale problems locally and constructing finite large-scale transfer coefficients. The method also introduces new basis function designs to ensure mass conservation when reconstructing the fine-scale velocity field. This method can reconstruct a velocity field that satisfies local fine-scale mass conservation, making the global solution at the fine scale more accurate. For the multiscale finite element method, it introduces multiple basis functions to enhance the model’s ability to handle complex problems. At the same time, spectral reduction techniques are employed to maintain model accuracy while reducing computational complexity. Building on this, the mixed Generalized Multiscale Finite Element Method\cite{chen2020generalized} (mixed GMsFEM) combines multiscale basis functions with direct velocity field solutions to ensure local conservation, particularly suitable for simulating high-contrast and multiphase flow. By selecting an appropriate offline basis function space, this method reduces the computational burden during online calculations while maintaining high accuracy using fewer basis functions.

\begin{figure}[h]
	\centering
	\includegraphics[width=1.0\linewidth]{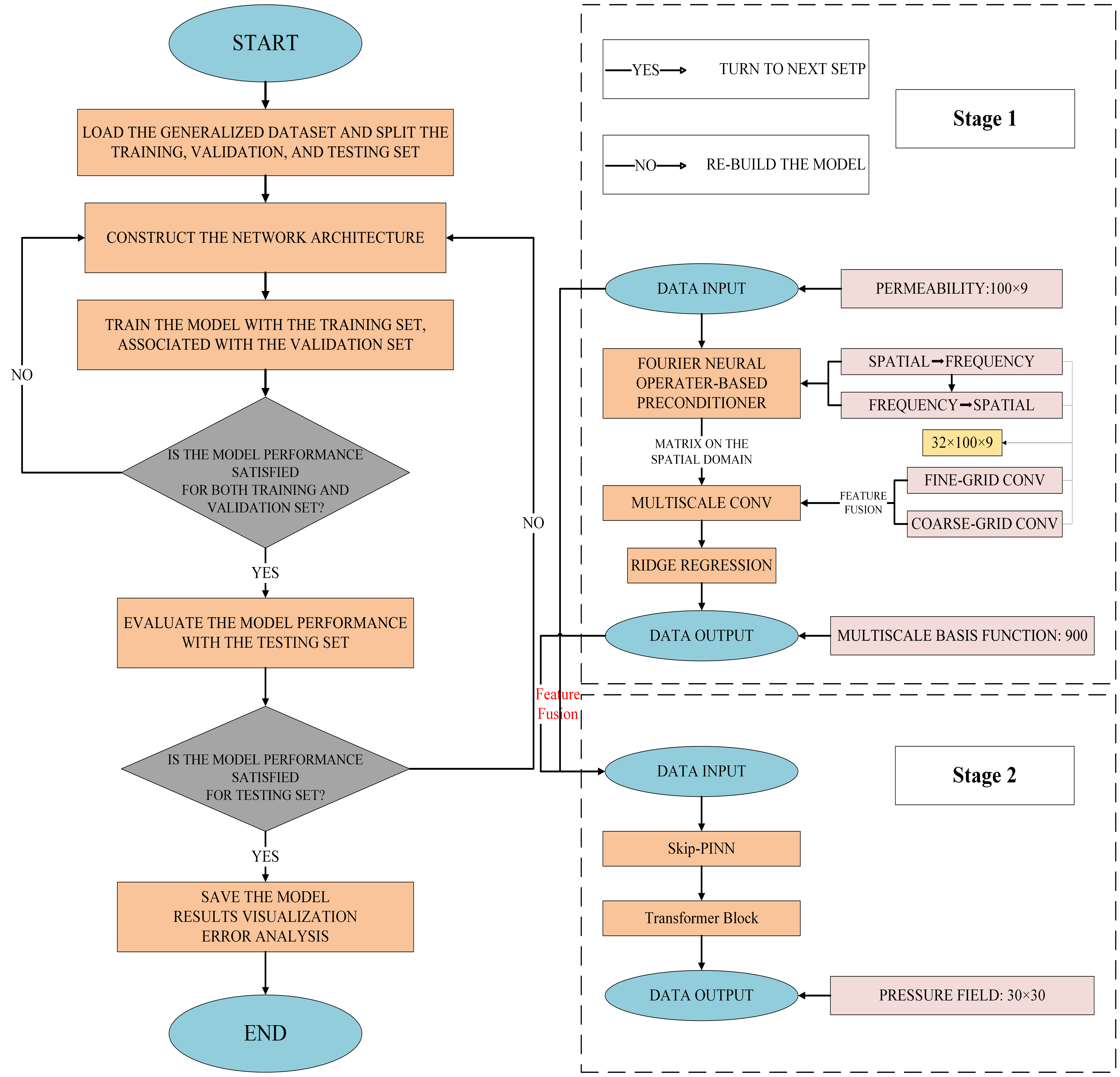}
	\caption{Flow chart of out research. Right: Total streamline of the two stages. Left: General architecture for each stage}
	\label{fig:flow chart}
\end{figure}

With the development of artificial intelligence (AI) technology, neural networks have demonstrated remarkable capabilities in learning complex mappings, and they exhibit computational speeds significantly faster than traditional numerical methods. Therefore, training neural network operators\cite{kovachki2023neural} to map random parameters to the solutions of PDEs has become an indispensable method in fields like scientific computing. The most common neural operators are finite operators, such as Convolutional Neural Networks (CNNs)\cite{raonic2024convolutional}, which extract features via local convolutional kernels. For instance, Choubineh et al.\cite{choubineh2022innovative} used a convolutional neural network to build a model for reconstructing multiscale basis functions and pressure field solutions in mixed GMsFEM. Through feature learning with convolutional operators, the model can map random parameters to the equation’s solution through complex linear and nonlinear transformations. Additionally, their deep ensemble model\cite{choubineh2023deep} integrates shortcut connections and linearly fits the results of different network paths to approximate the corresponding solution. This model achieves extremely high accuracy in solving the pressure field of Darcy equation. In contrast to finite operators, infinite operators do not discretize the input parameters and output solutions. A typical example of an infinite operator is the Fourier Neural Operator (FNO)\cite{li2020fourier,li2023fourier}, which uses Fourier transform to map the input function parameters from the spatial domain to the frequency domain, decomposing the information into multiple Fourier modes with different frequencies, and learning the weights of these modes in the frequency domain.

With the continuous advancement of deep learning\cite{wang2024recent}, Transformer\cite{vaswani2017attention} has received a lot of attention for their superior performance in processing sequence data. When solving porous media seepage problems, the self-attention mechanism of Transformer can capture complex dependencies in hydrodynamic behavior\cite{meng2023transformer,li2024transformer}, which is critical for simulating multi-phase flows in fracture networks. The PINNsFormer\cite{zhao2023pinnsformer} framework is proposed to combine Transformer's multihead attention mechanism with physical information neural network (PINN) to approximate the solution of PDE more precisely\cite{li2024physics}. The framework can effectively combine physical constraints with data-driven methods, thus improving the accuracy and efficiency in dealing with complex flow problems such as porous media seepage.

For the training of neural networks, there are two common methods: data-driven method and physics-informed method. In the data-driven approach, based on the data requirements of deep learning technology, traditional PDE solvers are usually used to simulate the solution functions of a large number of different parameter functions or fields, and then these data are used for the training of neural networks. While this training method can achieve desired results, it usually requires a large amount of data to train and can cause the trained operators to violate the laws of physics. In contrast, physics-informed methods\cite{pang2019fpinns,jagtap2020adaptive} ensure the superiority of local physical constraints and require fewer samples for training. Raissi et al.\cite{raissi2019physics} proposed the Physics-informed Neural Networks (PINNs), which embedded the physical information of PDE into the training process of the neural network through automatic differentiation technology. This technique is achieved by incorporating physical losses into the training objectives of the optimizer.

In this paper, a two-stage neural network operator is proposed to solve the Darcy flow problem step by step in the mixed GMsFEM framework: firstly, the multiscale basis function defined on the coarse grid is computed through data-driven model training, and then the pressure field is solved through physics-informed model training. \hyperref[fig:flow chart]{Fig.\ref{fig:flow chart}} shows the flow chart of our study. On the left is the process of building the entire model, which applies to two stages. The right side shows the general architecture of the two-stage model training.

The remainder of this paper will address different aspects of the study. \hyperref[sec:dataset]{Sec.\ref{sec:dataset}} will mainly explain the dataset we used; \hyperref[sec:stage1,sec:stage2]{Sec.\ref{sec:stage1} and \ref{sec:stage2}} will describe the work carried out in the two stages, including the methods used and the reconstruction tasks that need to be implemented. \hyperref[sec:results]{Sec.\ref{sec:results}} will explain the results and evaluation of the two-stage models. \hyperref[sec:discussion]{Sec.\ref{sec:discussion}} will summarize the main findings of this study and discuss its practical implications and future research directions. The appendix will provide additional details on the theoretical framework of the mixed GMsFEM method applied to our problem and present examples of our results. In the \hyperref[appendix:mixed gmsfem]{\ref{appendix:mixed gmsfem}}, we will introduce the knowledge of mixed GMsFEM for Darcy's equation. And the \hyperref[appendix:examples]{\ref{appedix:examples}} will illustrate some more examples of the results of our proposed method.

\section{Dataset Description}
\label{sec:dataset}

\begin{figure}[t]
	\centering
	\includegraphics[width=0.9\linewidth]{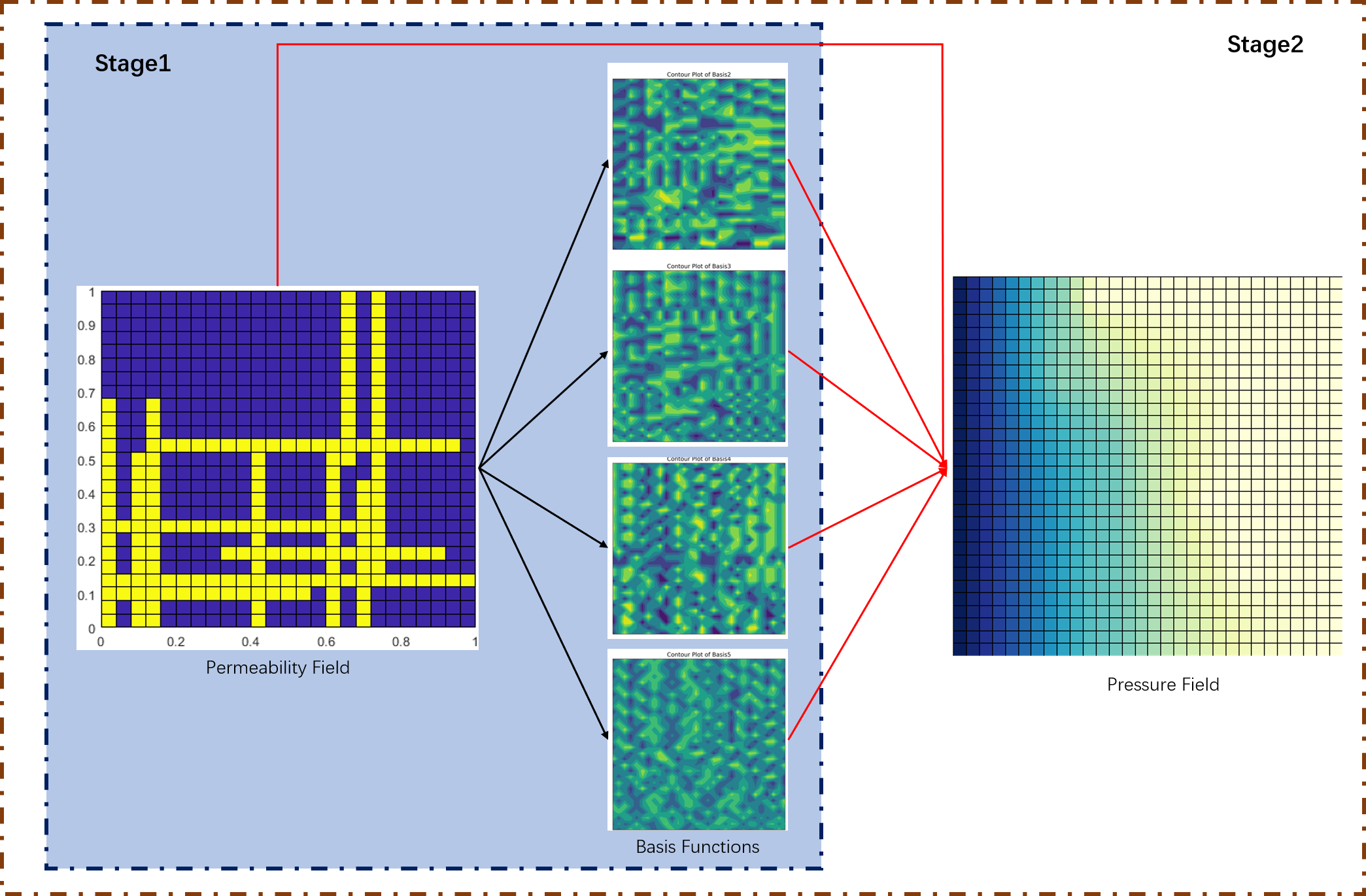}
	\caption{Dataset of our research. Permeability fields (left), multiscale basis functions (mid), and pressure fields (right).}
	\label{fig:dataset}
\end{figure}

\begin{figure}[h]
	\centering
	\includegraphics[width=0.2\linewidth]{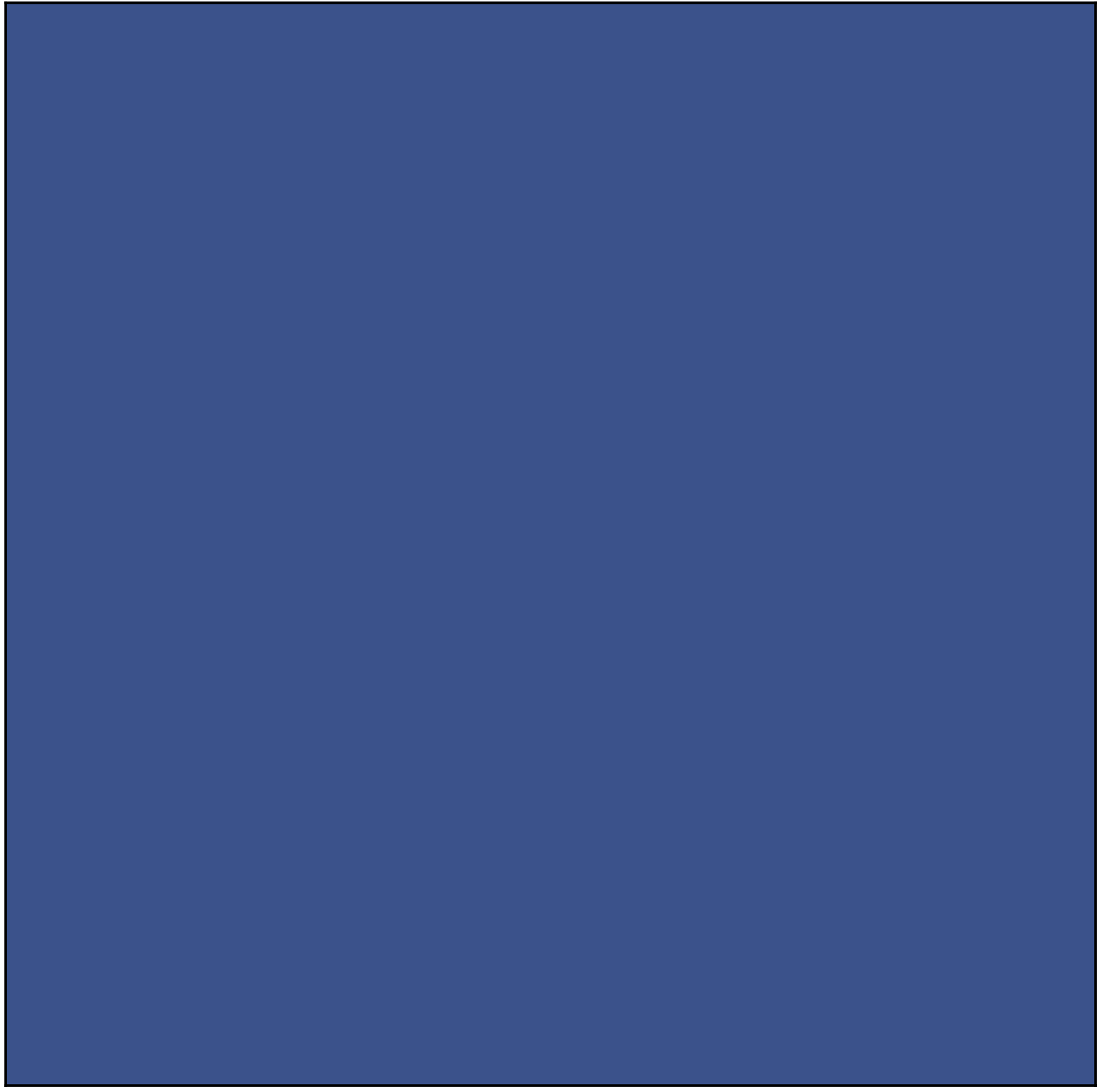}
	\caption{Contour plot of basis1. There is no value change in the contour plot. It is not related to the model training, as it is just a piecewise constant.}
	\label{fig:basis1}
\end{figure}

In the two-stage training process, we used data-driven and physics-informed methods, respectively. This means that in the first stage, a large amount of data is required for model training. In this stage, we parameterize random fields with specific covariance properties (high-contrast permeability fields) using Karhunen-Lo\`eve Expansion\cite{fukunaga1970application}. The computational domain is defined as $\Omega=[0,1]^2$, and the corresponding multiscale basis functions have values ranging from -1 to 1 ($-1 < \text{bf values} < 1$). The first basis function (basis1) is a piecewise constant, and we do not need to reconstruct this data using deep learning (see \hyperref[fig:basis1]{Fig.\ref{fig:basis1}}). Therefore, the multiscale basis functions involved in the training are basis2, basis3, basis4, and basis5.

\hyperref[fig:dataset]{Fig.\ref{fig:dataset}} shows an example of the dataset in our research, from left to right: high-contrast permeability field, reconstructed multiscale basis functions contour plot (reshaped to $30 \times 30$), and a pressure field. In a multiscale system, the fine grid is configured as a $30\times30$ homogenized grid, and the coarse grid is a $10\times10$ grid, that is, each coarse mesh contains $3\times3$ fine grid. In the grid system, the coarse grid may contain all or part of the fractures, which can be parallel or intersecting. In the parameter setting of permeability, the matrix permeability $K_m$ is between \{ 1, 2, 3, 4, 5 \} millidarcy values, the fracture permeability $K_f$ is between \{ 500, 750, 1000, 1250, 1500, 1750, 2000 \} millidarcy values, and the fracture number is between 1 and 25 random integers. The combination of three random parameters can generate 875 separate combinations for all for (permeability field, multiscale basis function) pairs. By iteratively generating these combinations, a total of 177,800 samples were generated. Considering the possibility of generating duplicate data, we removed 6,537 duplicates after removing the duplications. The dataset was divided into a 6:2:2 ratio of 102,757 training samples, 34,252 validation samples, and 34,254 test samples. The pressure field is a two-dimensional matrix obtained directly by mixed GMsFEM. For stage2, there are totally 1,700 samples envolved in physics-informed model training.

In the initial stage, the permeability field is represented as a $100\times 9$ two-dimensional vector, where 100 represents the number of coarse grids in each field and 9 represents the number of fine grids in each coarse grid. The four basis functions remain $900\times 1$ vectors. The pressure field is a two-dimensional matrix of $30\times 30$. Considering that batch normalization will be used in the convolutional network block, we do not normalize the permeability field.

\section{Multiscale Preconditioner for Reconstruction of Basis Functions: Stage 1}
\label{sec:stage1}
In this section, we will delve into the initial phase of the proposed method: initially, employing preconditioner techniques to refine the raw data; subsequently, leveraging multiscale deep learning strategies to extract pivotal features; and ultimately, reconstructing multiscale basis functions defined on the coarse grids.

\subsection{Domain Transform-based Feature Extraction}
The first step in this phase of the model involves using a Fourier Neural Operator-based neural network for the first stage feature extraction\cite{li2020fourier}, aimed at extracting frequency information inherent in the data and filtering redundant information. This part employs the Fourier integral operator $\mathcal{K}$ to transform the input $x$ into a spectral representation, a process that captures both local and global dependencies, thereby gaining insights into the spatial relationships within multiscale basis functions.

Define the Fourier integral operator
\begin{equation}
	\mathcal{K}(\phi)v_t = \mathcal{F}^{-1} \left( R_{\phi}\cdot (\mathcal{F}v_t)(x) \right),\ \ \forall x  \in D
	\label{Fourier intrgral operator}
\end{equation}
where $R_{\phi}$ is a Fourier periodic function parameterized by $\phi \in \Theta_{\mathcal{K}}$. The operator $\mathcal{K}$ plays a pivotal role in the domain transformation operation. The Fourier transform $\mathcal{F}$ will be utilized to transfer the input permeability field to the frequency domain:

\begin{align}
	(\mathcal{F}K)(\xi) & = \langle K, \psi \rangle_{L(D)} = \int_x v(x)\psi(x,\xi)\mu(x) \\
	\approx \sum_{x \in \mathcal{T}} v(x) \notag
\end{align}
here, $\psi(x,\xi)=\exp (2\pi i \langle x,\xi \rangle) \in L(D)$ is the Fourier basis function, $\xi$ represents the frequency mode used for nonlinear transformations within the network, indicating the Fourier modes to be retained. $\mathcal{T}$ is a uniform grid sampled from the distribution $\mu$.

\begin{figure}[t]
	\centering
	\includegraphics[width=1.0\linewidth]{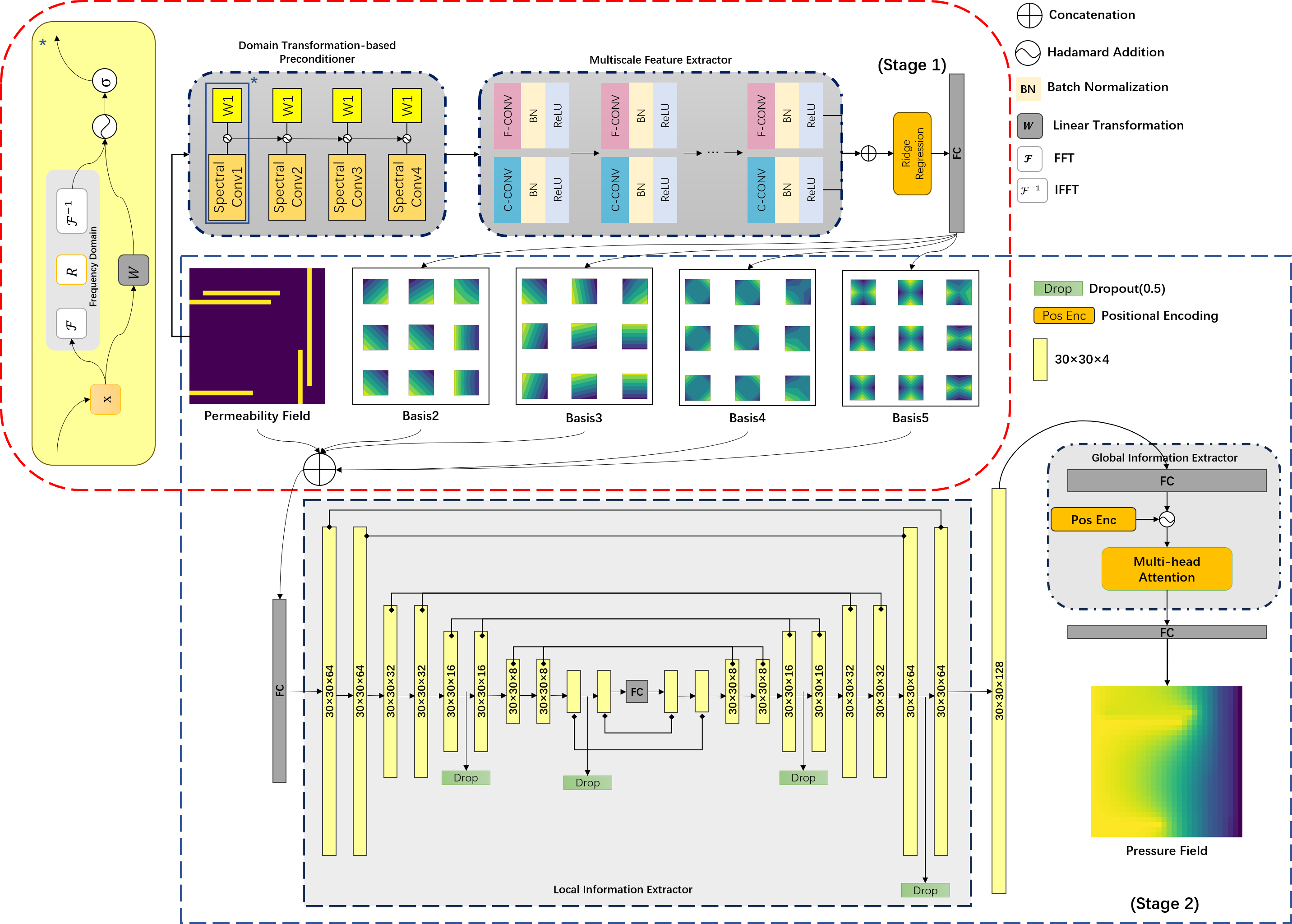}
	\caption{\textbf{Structure of total neural operator}. Data-driven model training for stage1 (Red): input feature (permeability field) $\mapsto$ output1 (multiscale basis functions 2-5). There are two main structures for this stage: Domain Transformation-based Preconditioner, and Hierarchical Multiscale Feature Extractor. Physics-informed model training for stage2 (Blue): input2 (permeability fields with 4 basis functions) $\mapsto$ output2 (pressure fields). Main sturctures for stage2: Local \& global information extractor.}
	\label{fig:Network structure}
\end{figure}

In our research, operator $\mathcal{F}$ is implemented via a two-dimensional Fast Fourier Transform (2D-FFT), marking the first critical step in the neural network. Specifically, the transformation is defined as:
\begin{equation}
	\hat{x}(\tilde{\xi})=\int_{\Omega} x\exp \left( -2\pi i \langle x, \tilde{\xi} \rangle \right)
\end{equation}
where $\Omega \in \left[ 0,1 \right]^2 \subset \mathbb{R}^2$ is the computational domain. After converting the data to the frequency domain, convolution operations are converted into element-wise multiplications, which can be regarded as linear transformations in the frequency domain. This transformation not only reveals the frequency characteristics of the data but also significantly enhances computational efficiency:
\begin{equation}
	\hat{y}(\tilde{\xi})=\tilde{\mathcal{K}}(\tilde{\xi})\cdot \hat{x}(\tilde{\xi})
\end{equation}
where $\tilde{\mathcal{K}}(\cdot)$ is the learnable kernel in the frequency domain, allowing the interactions under different spatial scales.

After performing spectral convolution, the weights of $R_{\phi}$ complete the necessary operations in the frequency domain. Subsequently, the data extracted after feature extraction will be mapped back to the spatial domain through the inverse Fourier transform (IFFT):
\begin{equation}
	y = \int_{\tilde{\xi}} \hat{y}\exp \left( 2\pi i \langle x, \tilde{\xi} \rangle \right) d\tilde{\xi}
\end{equation}

The aforementioned operations form our preconditioner. This architecture ensures that the model is trained in a feature space rich with information, capturing the required modes. Furthermore, it reduces the model complexity from $O(n^2)$ to $O(n\log n)$, enhancing convergence speed and enabling adaptation to multiscale and boundary conditions.

\subsection{Parallel Hierarchical Network for Multiscale Information}
In this section, we will introduce the second part of Stage 1. After the frequency-based learning, data will be fed into a CNN framework with a parallel architecture for further feature learning.

Although multiscale basis functions are defined on a coarser grid system, valuable information is still present in the finer grid system. To capture this information, we have designed convolutional filters of two different scales for information learning, which do not alter the size of the data. Thanks to the idea of oversampling, our multiscale network is implemented through two parallel convolutional modules. In this design, the output of the preprocessor will pass through a larger-scale network ($3 \times 3$) and a smaller-scale network ($1 \times 1$) simultaneously. The $1 \times 1$ convolutional filter does not involve complex data operations and can be considered an expansion of the original data channels.

Given the lightweight architecture of our designed neural network, a shallower depth may lead to issues such as overfitting. To avoid this, we use ridge regression to fit the outputs of the two CNN paths, effectively introducing $L2$ regularization into the model:
\begin{equation}
	y=W_{\text{Ridge}} \cdot x_{\text{flattened}} + b
\end{equation}
where $W_{Ridge}$ refers to the learnable weights. Additionally,  a regularization term is included in the loss function:
\begin{equation}
	\mathcal{L} = \frac{1}{N}\sum_{i=1}^{N} \left( y_i - \hat{y}_i \right)^2 + \lambda  \left \| W_{\text{Ridge}} \right \|_2^2
\end{equation}

The regularization term on the right side of the above equation helps prevent overfitting and enhances the model's generalization capability. The network structure for this stage is shown in \hyperref[fig:Network structure]{Fig.\ref{fig:Network structure}-stage1}.

\subsection{Evaluation Metircs for this stage}
Although we propose a two-stage approach, it is imperative to validate the experimental outcomes of the model from the first stage before proceeding to the second. At this juncture, our focus is particularly on the accuracy of the model's fit to the data. Consequently, we opt to evaluate the model using the Mean Squared Error ($\text{MSE}$) and the Coefficient of Determination ($\text{R}^2$).

$\text{MSE}$ quantifies the average squared difference between predicted and actual values, offering insights into overall prediction accuracy and penalizing larger errors more severely.
\begin{equation}
	\text{MSE}=\frac{1}{N} \sum_{i=1}^N \left(y_i - \hat{y}_i \right)^2
\end{equation}

$\text{R}^2$ represents the proportion of variance in the dependent variable that is explained by the independent variables, serving as an indicator of the model’s fit quality.
\begin{equation}
	\text{R}^2=1-\frac{\sum_{i=1}^N \left(y_i - \hat{y}_i \right)^2}{\sum_{i=1}^N \left( y_i - \bar{y} \right)^2}
	\label{eq:R2}
\end{equation}
where $y_i$, $\hat{y}_i$, $\bar{y}$ denote the true value, predicted value of the \textit{i}-th sample and average of true value.

\newpage
\section{Physical Connected Operator Solving Pressure Fields: Stage 2}
\label{sec:stage2}
This section will detail the second phase of our methodology: a PINN-Transformer architecture capable of reconstructing pressure fields from small sample data, featuring residual connections. In this architecture, the permeability field $\kappa$ and the multiscale basis functions $bf$ obtained from the first stage will serve as input data to learn the pressure solution $p(\kappa,bf)$ of Darcy equation. In other words, we aim to develop a parameterized deep learning model $\mathcal{G}$ that facilitates the mapping from input to output:
\begin{equation}
	\mathcal{G}:\ \mathbb{R}^{2\times 2} \mapsto \mathbb{R}^2,\ \ \ (\kappa,bf) \mapsto p_{\theta}(\kappa,bf)
\end{equation}
this mapping can be considered as a differentiable mapping from the parameter space $\mathbb{R}^M$ to the function space $\mathcal{V}:\ \theta \mapsto p_{\theta} \in \mathcal{V}$. 

\subsection{Physics-informed Operator with Non-neighbored Accelerator}
In contrast to traditional CNN architectures, we aim to develop a mapping operator that not only achieves high accuracy but also strictly adheres to the laws of physics, particularly in the context of Darcy flow modeling, where it exhibits a certain level of interpretability. To accomplish this, we have employed Physics-Guided Deep Learning Methods (PGDL Methods).

PGDL methods combine the powerful nonlinear mapping capabilities of deep learning with prior knowledge of physical laws, thereby significantly enhancing the accuracy, stability, and interpretability of predictive models. The PINN serves as a specific implementation of PGDL, embedding physical information into the model through residual learning.

Define the residual of Darcy equation:
\begin{align}
	\mathcal{R}(x;\theta) &= - \nabla \cdot \left( \kappa(x)\nabla p_{\theta}(x) \right)-f(x) \\
	&= -\left( \nabla \kappa(x) \cdot \nabla p_{\theta}(x) + \kappa(x) \Delta p_{\theta}(x) \right) - f(x)
\end{align}
where $\nabla p_{\theta}(x)$ is a d-dimensional vector with components $\frac{\partial p_{\theta}}{\partial x_i}(x)$, $\Delta = \sum_{i=1}^d \frac{\partial^2}{\partial x_i^2}$ is the Laplacian operator, $\theta$ represents the parameters of the neural network, and $f(x)$ is the known source term.

To embed physical residuals into a neural network, we utilize a loss function to achieve this objective. Traditional neural networks typically employ the MSE loss for regression tasks or the Binary Cross-Entropy (BCE) loss for binary classification tasks. However, in our research, we need to incorporate physical information into the loss function to ensure that the model's predictions adhere to the laws of physics. Additionally, we must consider the loss associated with the data itself to enhance the model's fit to the actual observed data.

The data loss term $\mathcal{L}_{data}$ is derived from measured data points of known boundary data, consistent with the MSE Loss:
\begin{equation}
	\mathcal{L}_{\text{data}}(\theta) = \frac{1}{N}\sum_{j=1}^N \| p_{\theta}(x_j)- p_{\text{true}}(x_j) \|^2
\end{equation}
where $N$ is the number of sample points, and $p_{\text{true}}(\cdot)$ denotes the true data of the pressure field. The gradient of the data term loss is given by:
\begin{equation}
	\nabla_{\theta} \mathcal{L}_{\text{data}}(\theta) = \frac{2}{N}\sum_j \left[ 
	p_{\theta}(x_j)- p_{\text{true}}(x_j) \right] \nabla_{\theta} p_{\theta}(x_j)
\end{equation}

The physical residual loss is defined as the MSE of the physical residuals, with the formula:
\begin{equation}
	\mathcal{L}_{\text{physics}} = \frac{1}{N_r}\sum_{i=1}^{N_r} \| \mathcal{R}(x_i ; \theta) \|^2
\end{equation}
here, $N_r$ represents the number of sampling points within the domain. The gradient of the physical residual loss is expressed as:
\begin{align}
	\nabla_{\theta} \mathcal{L}_{\text{physics}}(\theta) &= \frac{2}{N_r} \sum_i \mathcal{R}(x_i ; \theta) \nabla_{\theta}\mathcal{R}(x_i ; \theta), \\ 
	\nabla_{\theta}\mathcal{R}(x_i ; \theta)&=-k \nabla_{\theta}(\Delta p_{\theta}(x_i)) \notag
\end{align}

\hyperref[fig:Network structure]{Fig.\ref{fig:Network structure}-stage2-local information extractor} illustrates the operator architecture we propose, which consists of symmetrically distributed V-shaped convolutional layers and embeds physical information through residual learning during backpropagation. The increase in model complexity may lead to a higher risk of overfitting, which is undesirable. To address this, in addition to incorporating Dropout layers to mitigate the potential for overfitting, we have also introduced L2 regularization into the model, which adds a regularization term to the loss function:
\begin{equation}
	\lambda_{L2}\| \theta \|_2^2 = \lambda_{L2} \sum_{\theta_k \in \theta}\theta_k^2
\end{equation}
and its gradient:
\begin{equation}
	\nabla_{\theta} \left( \lambda_{L2}\| \theta \|_2^2 \right) = 2 \lambda_{L2}\theta
\end{equation}

Taking into account the relative importance of data loss and physical loss within the model, we have designed the total loss function and the total gradient as follows:
\begin{align}
	\mathcal{L}(\theta) &= \alpha \mathcal{L}_{\text{data}}(\theta)+\beta \mathcal{L}_{\text{physics}}(\theta) + \lambda_{L2}\| \theta \|_2^2 \notag \\
	\nabla_{\theta}\mathcal{L} &= \alpha\nabla_{\theta}\mathcal{L}_{\text{data}} + \beta \nabla_{\theta} \mathcal{L}_{\text{physics}} + 2\lambda_{L2}\theta
\end{align}
where $\alpha$ and $\beta$ are the weights for the data loss and physical loss, respectively, and satisfy $\alpha + \beta =1$, while $\lambda_{L2}$ is the regularization coefficient.

As the network architecture becomes increasingly complex, a significant loss of data features may occur during the transmission of information between layers, potentially leading to the gradient vanishing problem. Additionally, a large number of parameters can substantially prolong the training time of the model. To effectively expedite the training process, we have incorporated accelerators based on residual connections into the network architecture, aiming to mitigate these issues and enhance training efficiency.

\begin{figure}[h]
	\centering
	\includegraphics[width=1.0\linewidth]{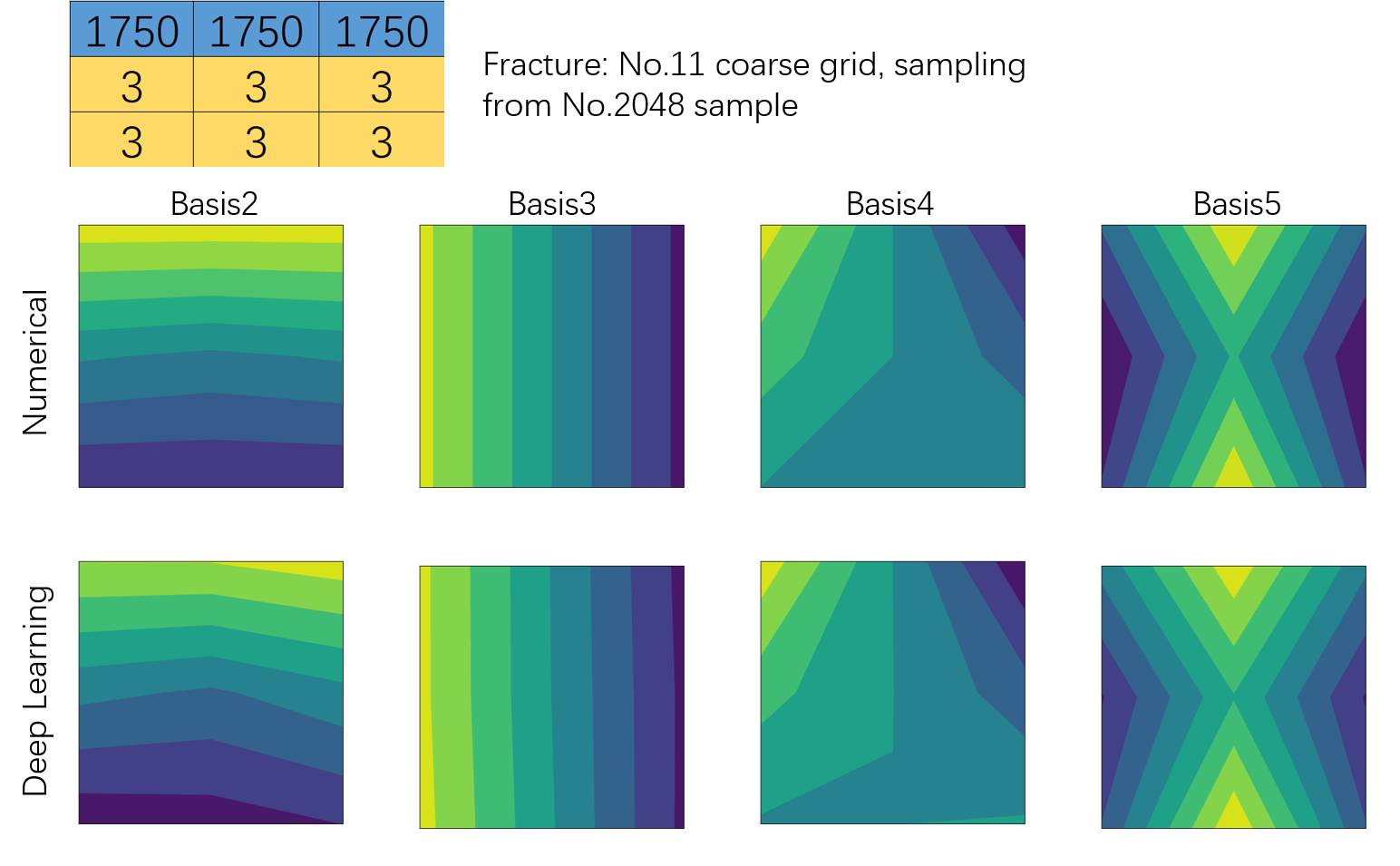}
	\caption{Example of Reconstruction for basis functions, sampling from No.11 coarse grid of No.2048 validation sample (with fractures).}
	\label{fig:val fracture}
\end{figure}

\begin{figure}[h]
	\centering
	\includegraphics[width=1.0\linewidth]{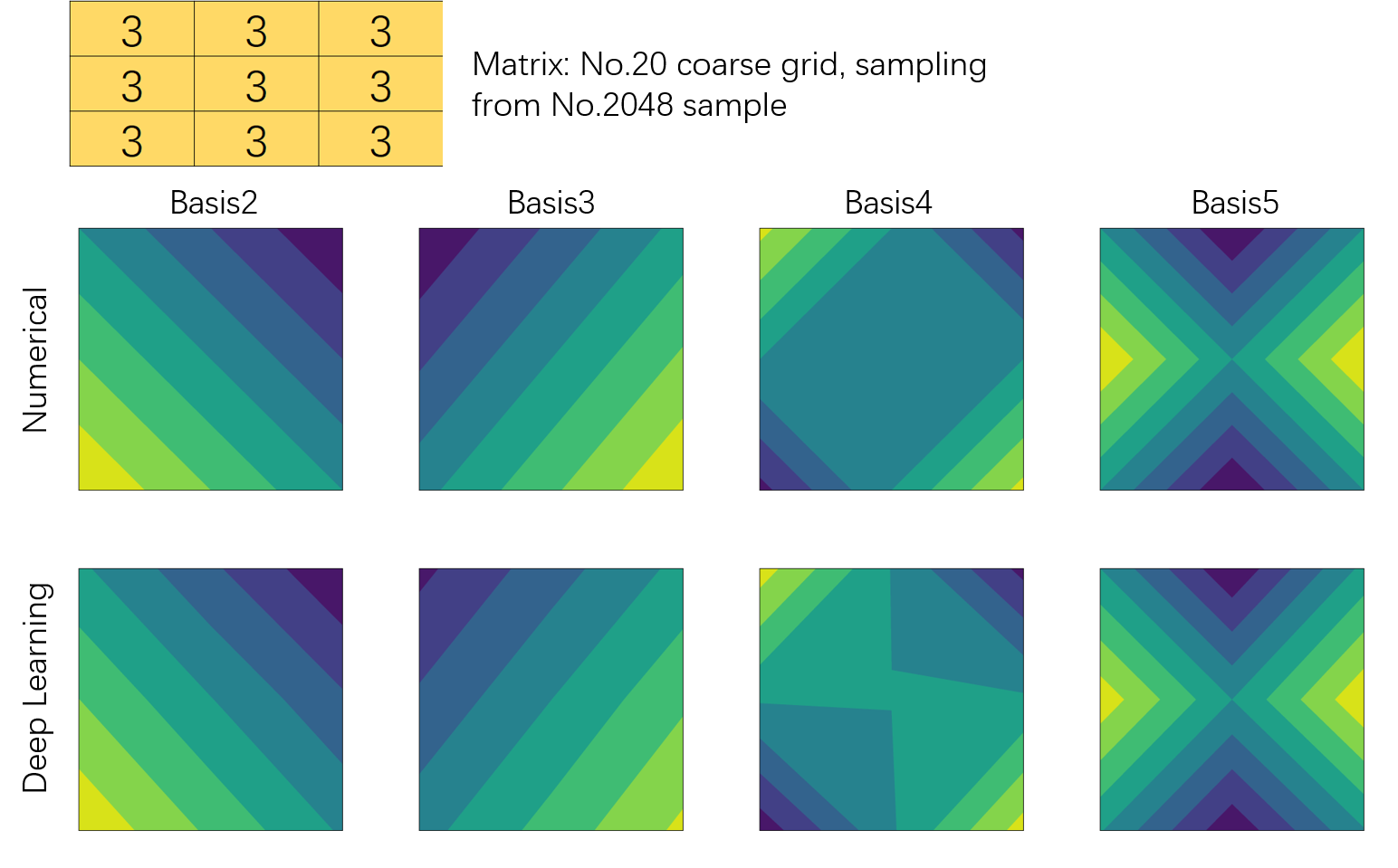}
	\caption{Example of Reconstruction for basis functions, sampling from No.20 coarse grid of No.2048 validation sample (without fractures).}
	\label{fig:val matrix}
\end{figure}

Consider skip connections between non-adjacent layers, which establish direct links between different hierarchical levels. Specifically, we introduce a mechanism that allows the output $h^{(a)}$ from layer $a$ to directly contribute to the output $h^{(b)}$ of layer $b$:
\begin{equation}
	h^{(b)}=h^{(a)}+\mathcal{G}^{(b)}(h^{{(b-1)}},\theta^{(b)})
\end{equation}
where $\mathcal{G}^{(b)}$ represents the original mapping function from layer $(b-1)$ to layer $b$, a process that is independent of the skip connection.

These non-adjacent skip connections act as shortcuts from earlier layers to later layers in the computational graph, creating a multipath structure. During forward propagation, features from layer $a$ can directly influence the output of layer $b$ without undergoing the nonlinear transformations of the intermediate layers. This design ensures that the feature information extracted by the early layers is not easily diminished or lost during deep propagation.

Assume that the existence of a skip connection in the network that directly jumps from the input layer to the \textit{L}-th layer:
\begin{equation}
	h^{(L)}=h^{(0)}+\sum_{\text{paths}}\mathcal{H}_{\text{path}}(x;\theta)
\end{equation}
where $\mathcal{H}_{\text{path}}$ denotes the general transmission path from input to output, excluding the skip connection. During the training process of the neural network, backpropagation of gradients is essential, necessitating the solution for $\nabla_{h^{(0)}}h^{(L)}$:
\begin{equation}
	\frac{\partial h^{(L)}}{\partial h^{(0)}}=I\  + \sum_{\text{paths}} \frac{\partial \mathcal{H}_{\text{path}}}{\partial h^{(0)}}
\end{equation}

In the absence of skip connections or with only adjacent skip connections, $\mathcal{H}_{\text{path}}$ represents a deep composition of a series of mappings, which may lead to a decrease in gradient magnitude across layers. However, if non-adjacent paths exist, these paths can contribute at least one identity mapping $I$ when gradients are calculated using the chain rule. Even with other complex path Jacobian multiplications, this skip path ensures that the gradient includes a channel that has not undergone multiple nonlinear compressions:
\begin{equation}
	\frac{\partial h^{(b)}}{\partial h^{(a)}} = I \  + \cdots
\end{equation}

The $\cdots$ indicates the influence of other paths on $h^{(a)}$. This ensures that under the multilayer nonlinear transformation conditions of an MLP architecture, the gradient does not entirely rely on the multiplication of deeper layers, thus preventing rapid decay of the norm. Residual information can stably trace back to shallower layers during backpropagation, alleviating the vanishing gradient problem and accelerating model convergence.

The inclusion of skip connections provides a multilevel support path for gradient and information flow, reducing the difficulty of optimization. With long-range skip connections, the gradient signal is stronger and more stable during parameter updates, enabling the optimizer to quickly find favorable parameter update directions, thereby accelerating convergence.

By embedding physical information and residual connections into the neural network architecture, the neural network is compelled to adhere to the laws of physics, maintain effective information transfer under conditions of high model complexity, and expedite model training.

\subsection{Hierarchical Operator to Capture Global Information}
Although the aforementioned model, by embedding physical laws and residual connections into a deep learning framework, can effectively approximate the solutions to PDEs, its primary convolutional structure, which focuses on local features, limits its ability to capture global interactions and long-range dependencies. To address this limitation, we have integrated a Transformer module. The Transformer enhances the model's ability to capture global features and long-range dependencies through self-attention mechanisms and positional encoding, thereby improving the accuracy of predicting the pressure field $p_{\theta}(\cdot)$.

Initially applied to natural language processing tasks, the Transformer has gained widespread attention for its efficiency in capturing dependencies within sequences. We have connected the Transformer module to the output end of the PINN, aiming to enhance the model's ability to capture global information and long-range dependencies through global self-attention mechanisms and positional encoding architectures. The overall process is as follows:
\begin{equation}
	p_\theta(x) \xrightarrow{\scriptstyle \text{(Flatten)}} z \xrightarrow{\scriptstyle \text{(Pos Enc)}} z' \xrightarrow{\scriptstyle \text{(Transformer)}} z'' \xrightarrow{\scriptstyle \text{(FC)}} \hat{p}_\theta(x) \notag
\end{equation}

Self-attention mechanism is at the core of the Transformer architecture, enabling the model to dynamically weigh the relevance between different parts of the data. For each token in the output sequence, the self-attention mechanism calculates its relevance weights with all other tokens and performs a weighted sum of the values based on these weights, achieving global information integration. Given the input sequence $z'=[z_1', z_2', ...,z_N']$, we project it linearly as follows:
\begin{equation}
	Q=z'W^Q,\ \ K=z'W^K,\ \  V=z'W^V \notag
\end{equation}
here, $W^Q$, $W^K$, $W^V \  \in \mathbb{R}^{d \times d_k}$ are learnable matrices used for the linear transformations of Queries, Keys, and Values, respectively.

Self-attention is implemented through scaled dot-product attention:
\begin{equation}
	\text{Attention}(Q,K,V)=\text{softmax} \left( \frac{QK^T}{\sqrt{d_k}} \right)V
\end{equation}
where $\sqrt{d_k}$ is a scaling factor to prevent the dot product from becoming too large.

The multi-head attention extends the single self-attention mechanism by computing multiple different attention heads in parallel and concatenating their results to obtain a more expressive output. Specifically, the calculation of multiple attention heads is as follows:
\begin{equation}
	\text{Multihead}(Q,K,V)=\text{Concat}(\text{head}_1,\text{head}_2,...,\text{head}_h)W^O
\end{equation}
each $\text{head}_i$ is computed as $\text{Attention}(Q_i,K_i,V_i)$, processing different linear projection subspaces, and $W^O$ is the output projection matrix belonging to $\mathbb{R}^{h\cdot d_k \times d}$, where $h$ is the number of attention heads, $d_k$ is the dimension of the key and query vectors, and $d$ is the dimension of the output.

\begin{figure}[h]
	\centering
	\includegraphics[width=1.0\linewidth]{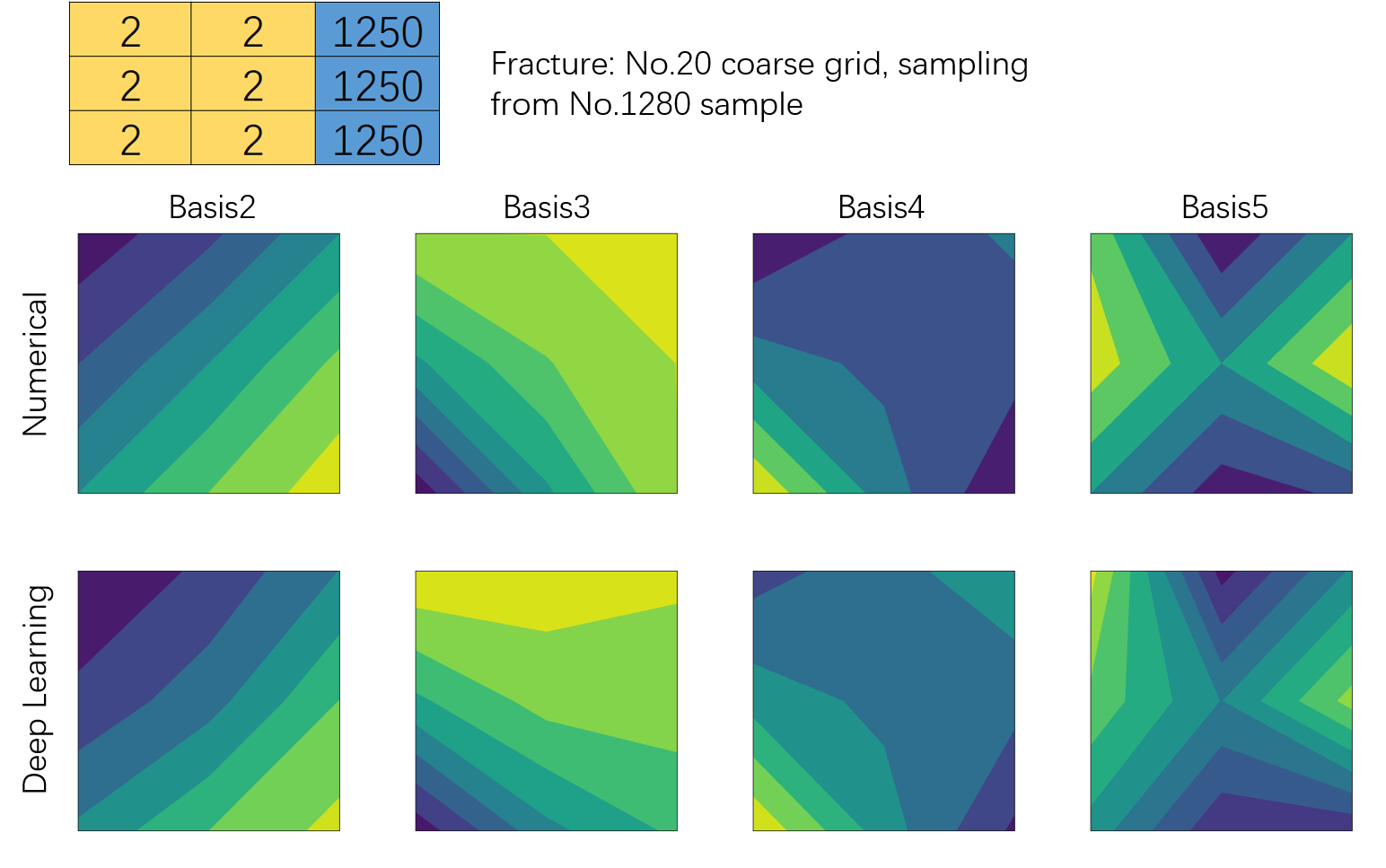}
	\caption{Example of Reconstruction for basis functions, sampling from No.20 coarse grid of No.1280 testing sample (with fractures).}
	\label{fig:test fracture}
\end{figure}

The architecture composed solely of multi-head attention is insensitive to the order of the sequence. To introduce spatial positional information, positional encodings are added to the input features of the module. Positional encodings provide each sequence token with positional information relative to its location in space, allowing the Transformer to recognize and utilize spatial relationships.

The definition of positional encoding is as follows:
\begin{equation}
	\text{PE}(i, 2k) = \sin\left(\frac{i}{10000^{\frac{2k}{d}}}\right), \quad \text{PE}(i, 2k+1) = \cos\left(\frac{i}{10000^{\frac{2k}{d}}}\right) \notag
\end{equation}
where $i$ is the position of the token in the sequence, $k$ is the dimension index, and $d$ is the total dimension of the embeddings. The 10,000 in the denominator is simply a number large enough to ensure that the positional encoding does not duplicate. The use of sine and cosine functions with different frequencies allows the model to distinguish between different positions and dimensions within the sequence. The Positional encoding is added to the input embedding $z_i$ through
\begin{equation}
	z_i'=z_i + \text{PE}(i)
\end{equation}
where $z_i'$ represents the input embedding after the addition of positional encoding. By integrating positional encoding into the output of the convolutional network, the model ensures that the Transformer architecture can identify the exact spatial location of each feature. This allows the model to fully utilize spatial information during the self-attention calculation, significantly enhancing its ability to capture global information and long-range dependencies.

In the context of solving Darcy Flow problem, this capability is particularly important. Certain parts of the pressure field in Darcy Flow may be influenced by the distribution of permeability and source terms at a distance. Therefore, the architecture of the Transformer model plays a significant positive role in understanding and predicting these complex spatial relationships.

\begin{figure}[h]
	\centering
	\includegraphics[width=1.0\linewidth]{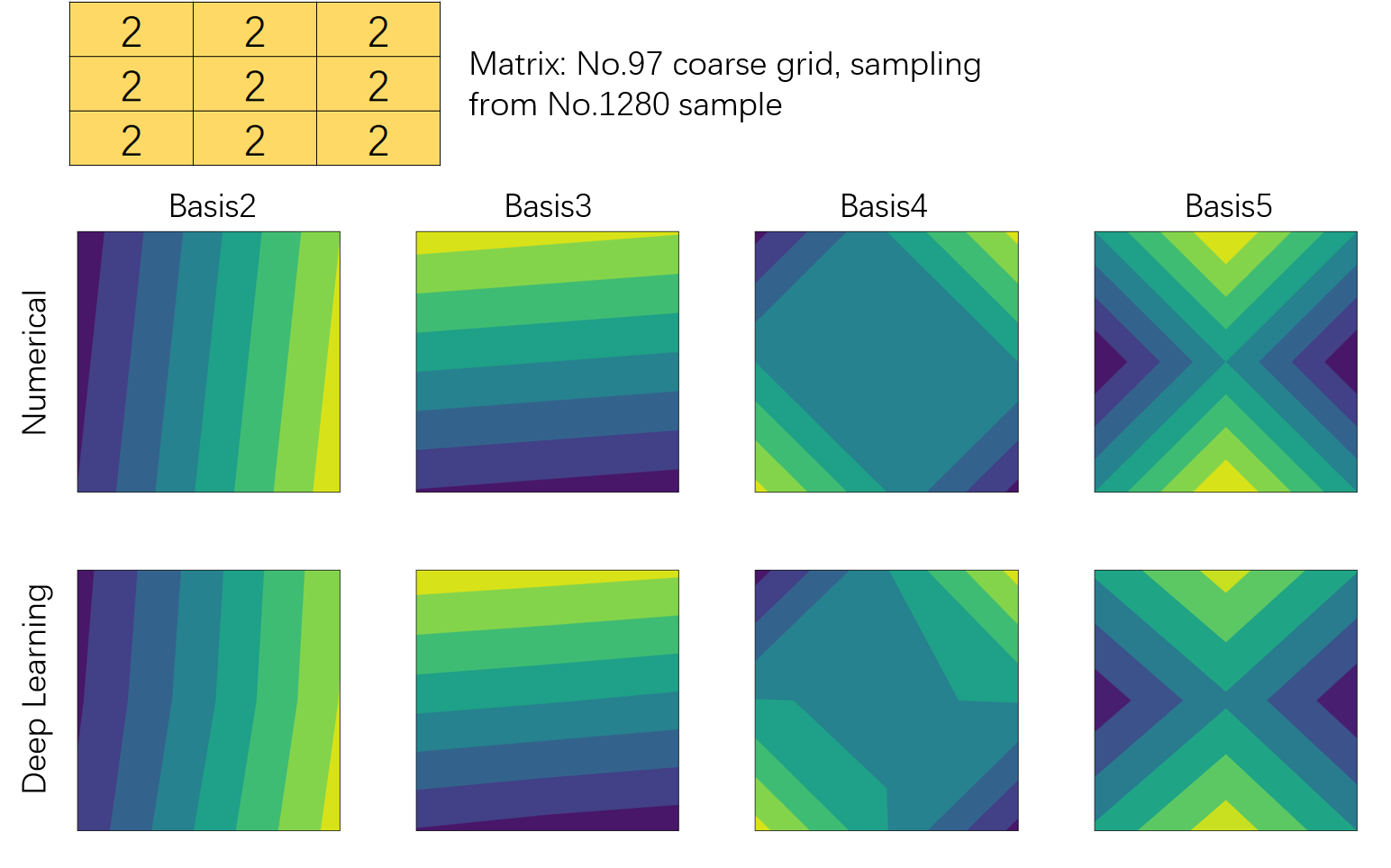}
	\caption{Example of Reconstruction for basis functions, sampling from No.97 coarse grid of No.1280 testing sample (without fractures).}
	\label{fig:test matrix}
\end{figure}

\subsection{Evaluation Metircs for this stage}
Since we add physical information to the loss function, it is not appropriate to use the mean square error as an evaluation index directly. To do this, we use the $\text{R}^2$ and define the average absolute physical residual (AAPR) to evaluate the coefficient of variation of the model and the degree of fit to the data and physical information, which has been demonstrated by \hyperref[eq:R2]{Equation.\ref{eq:R2}}.

Physics residual measures the bias of model prediction and physical constraints. To ensure our model can not only fit the data, but also satisfy the physical law, we computed the physical residual, and evaluated the physical consistency of the model using AAPR:
\begin{equation}
	\text{AAPR}=\frac{1}{N}\sum_{i=1}^N |\mathcal{R}(x_i,p_i)|
\end{equation}
where $\mathcal{R}(x_i,p_i)$ os the physical residual of \textit{i}-th sample. The smaller the AAPR, the more consistent the model’s predicted pressure field is with physical constraints, indicating better performance in terms of physical consistency. Therefore, AAPR is an important metric for measuring the degree of satisfaction of physical model constraints.

In this section, we propose a neural operator that integrates a convolutional operator with physical residual links and a Transformer module, designed to predict the pressure field from multi-scale basis functions and permeability fields. We define $\mathcal{G}_{\text{PINN}}$ as the first part of the mapping in Stage 2, mathematically expressed as:
\begin{equation}
	\mathcal{G}_{\text{PINN}}:\  \mathbb{R}^{C\times H\times W} \times \mathbb{R}^M \mapsto \mathbb{R}^{N \times d}
\end{equation}
where $C$ represents the number of channels, $H$ and $W$ denote height and width, respectively, $M$ is the dimension of the input parameters, $N$ is the number of output features, and $d$ is the dimension of the features.

The Transformer module $\mathcal{G}_{\text{Trans}}$, acting as a high-dimensional operator, processes the output features of PINN, expressed as:
\begin{equation}
	\mathcal{G}_{\text{Trans}}:\  \mathbb{R}^{N\times d} \mapsto \mathbb{R}^{N\times d}
\end{equation}

The overall mapping relationship is represented as:
\begin{equation}
	\mathcal{G}_{\text{Total}} = \mathcal{G}_{\text{Trans}} \circ \mathcal{G}_{\text{PINN}}: \mathbb{R}^{C \times H \times W} \times \mathbb{R}^M \mapsto \mathbb{R}^{N \times d}
\end{equation}

The final predicted pressure field $\hat{p}(\theta)(\cdot)$ is given by:
\begin{equation}
	\hat{p}_\theta(x) = \mathcal{G}_{\text{Total}}(X) = \mathcal{G}_{\text{Trans}}\left(\mathcal{G}_{\text{PINN}}(x)\right) = \int_{\Omega} \mathcal{K}\left(x, x'\right) p_\theta\left(x'\right) dx'
\end{equation}

\section{Numerical Results}
\label{sec:results}
In this section we will demonstrate our results in two ways: evaluation metrics and data visualization.

For the first stage, reconstruction for 4 multiscale basis functions, we used MSE and R$^2$ to evaluate the model performance (as shown in \hyperref[table:basis functions metrics]{Tab.\ref{table:basis functions metrics}}). 

\begin{table}[h]
	\centering
	\caption{Evaluation metrics for different basis functions.}
	\label{table:basis functions metrics}
	\renewcommand{\arraystretch}{1.2} 
	\setlength{\tabcolsep}{8pt} 
	\begin{tabular}{c|c|c|c|c|c|c|c|c}
		\hline
		\multirow{2}{*}{} & \multicolumn{4}{c|}{MSE} & \multicolumn{4}{c}{$\text{R}^2$} \\ \cline{2-9} 
		& Training   & Validation   & Testing   & Total   & Training   & Validation   & Testing   & Total   \\ \hline
		Basis 2           & 0.0080     & 0.0155       & 0.0036    & 0.0086  & 0.9566     & 0.9310       & 0.9716    & 0.9537  \\ \hline
		Basis 3           & 0.0005     & 0.0053       & 0.0007    & 0.0017  & 0.9607     & 0.9007       & 0.9148    & 0.9402  \\ \hline
		Basis 4           & 0.0009     & 0.0011       & 0.0021    & 0.0012  & 0.9454     & 0.9357       & 0.9366    & 0.9417  \\ \hline
		Basis 5           & 0.0012     & 0.0014       & 0.0027    & 0.0016  & 0.9706     & 0.9642       & 0.9542    & 0.9659  \\ \hline
	\end{tabular}
\end{table}

As shown in the data presented in the table, the MSE of our proposed first-stage model on the test set for different basis functions is 0.0036, 0.0007, 0.0021, and 0.0027, respectively, while the corresponding R$^2$ values are 0.9716, 0.9148, 0.9366, and 0.9542. These values indicate that such performance is acceptable for deep learning models. To further enhance the credibility of our results, we selected samples from the dataset used in this stage, specifically choosing coarse grids that include fractures and those without fractures, and reconstructed the corresponding basis functions for visualization. These results can be observed in the following figures: validation set—\hyperref[fig:val fracture, val matrix]{Fig.\ref{fig:val fracture} and \ref{fig:val matrix}}; testing set—\hyperref[fig:test fracture, test matrix]{Fig.\ref{fig:test fracture} and \ref{fig:test matrix}}; training set—\hyperref[fig:train fracture, train matrix]{Fig.\ref{fig:train fracture} and \ref{fig:train matrix}}. It should be noted, however, that deep learning techniques cannot guarantee perfect learning of features from the data. Therefore, there is a possibility that the selected samples for visualization may include coarse grids with suboptimal reconstruction quality.

\begin{table}[h]
	\centering
	\renewcommand{\arraystretch}{1.2} 
	\setlength{\tabcolsep}{9pt} 
	\caption{R$^2$ Performance of Different Basis Function Combinations.}
	\begin{tabular}{l|c|c|c|c}
		\hline
		\textbf{Basis Functions} & \textbf{Training} & \textbf{Validation} & \textbf{Testing} & \textbf{Total} \\ \hline
		Basis 3 & 0.9960 & 0.9395 & 0.9068 & 0.9870 \\ \hline
		Basis 4 & 0.9965 & 0.9412 & 0.9111 & 0.9878 \\ \hline
		Basis 1+2 & 0.9942 & 0.9201 & 0.9032 & 0.9847 \\ \hline
		Basis 2+3 & 0.9945 & 0.9320 & 0.9064 & 0.9852 \\ \hline
		Basis 2+4+5 & 0.9959 & 0.9374 & 0.9093 & 0.9869 \\ \hline
		Basis 2+3+4+5 & 0.9960 & 0.9383 & 0.9018 & 0.9866 \\ \hline
	\end{tabular}
	\label{tab:r2_performance}
\end{table}

For the second stage, reconstruction for the pressure field, the same step to illustrate the results. \hyperref[r2_performance, aapr_performance]{Tab.\ref{tab:r2_performance} and \ref{tab:aapr_performance}} present the numerical results of the evaluation metrics used in our study. In this experiment, we combined various basis functions (including Basis 1, which requires no training) with their corresponding permeability fields and input them into the model for training. For our experiments, we randomly selected different numbers and combinations of basis functions for training. From the test set column in Tab.\ref{tab:r2_performance}, it can be observed that the R$^2$ values of our model are all greater than 0.9, indicating that the model performs excellently in fitting the pressure data. Furthermore, as shown in Tab.\ref{tab:aapr_performance}, the AAPR values are all at the magnitude of $1 \times 10^{-4}$ or smaller, strongly demonstrating that the proposed model not only accurately reconstructs the data but also adheres to the underlying physical laws of real-world phenomena.

\begin{table}[h]
	\centering
	\renewcommand{\arraystretch}{1.2} 
	\setlength{\tabcolsep}{9pt} 
	\caption{AAPR Performance of Different Basis Function Combinations.}
	\label{tab:aapr_performance}
	\begin{tabular}{l|c|c|c|c}
		\hline
		\textbf{Basis Functions} & \textbf{Training} & \textbf{Validation} & \textbf{Testing} & \textbf{Total} \\ \hline
		Basis 3 & \( 5.1871 \times 10^{-5} \) & \( 1.3509 \times 10^{-4} \) & \( 6.8263 \times 10^{-6} \) & \( 3.7421 \times 10^{-5} \) \\ \hline
		Basis 4 & \( 1.1654 \times 10^{-4} \) & \( 1.6001 \times 10^{-4} \) & \( 5.0488 \times 10^{-4} \) & \( 1.4194 \times 10^{-4} \) \\ \hline
		Basis 1+2 & \( 7.4166 \times 10^{-5} \) & \( 1.0020 \times 10^{-4} \) & \( 6.0759 \times 10^{-4} \) & \( 2.9245 \times 10^{-5} \) \\ \hline
		Basis 2+3 & \( 2.4464 \times 10^{-4} \) & \( 4.6596 \times 10^{-4} \) & \( 3.3928 \times 10^{-4} \) & \( 2.6323 \times 10^{-4} \) \\ \hline
		Basis 2+4+5 & \( 1.6015 \times 10^{-4} \) & \( 1.6996 \times 10^{-4} \) & \( 2.3803 \times 10^{-4} \) & \( 1.6531 \times 10^{-4} \) \\ \hline
		Basis 2+3+4+5 & \( 1.5022 \times 10^{-4} \) & \( 1.5124 \times 10^{-4} \) & \( 3.4182 \times 10^{-4} \) & \( 1.2134 \times 10^{-4} \) \\ \hline
	\end{tabular}

\end{table}

\begin{figure}[h]
	\centering
	\includegraphics[width=1.0\linewidth]{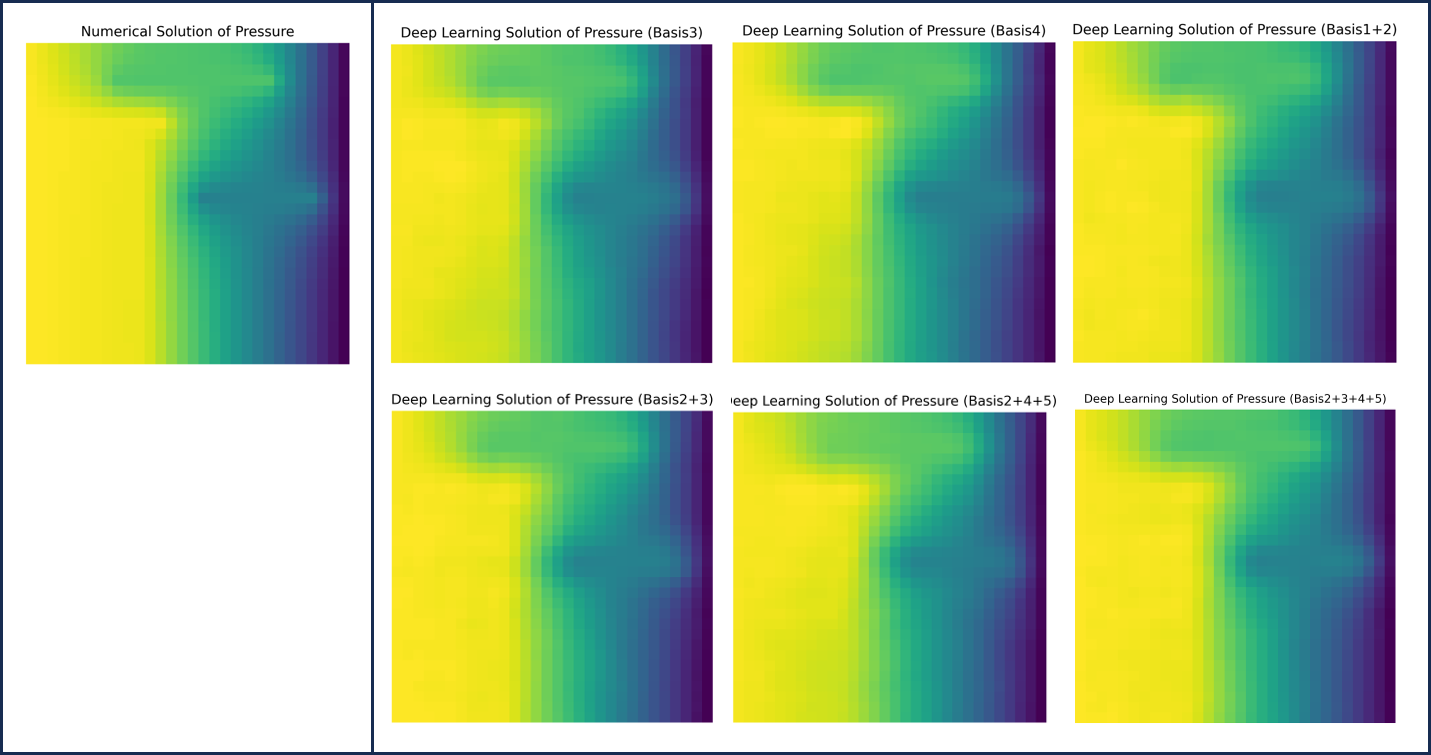}
	\caption{Comparison of the reconstruction results using different basis functions combination. Right: numerical solution. Left: deep learning solutions of different combinations. (No.1300 sample from the dataset)}
	\label{fig:results_comparison}
\end{figure}

\begin{figure}[h]
	\centering
	\includegraphics[width=1.0\linewidth]{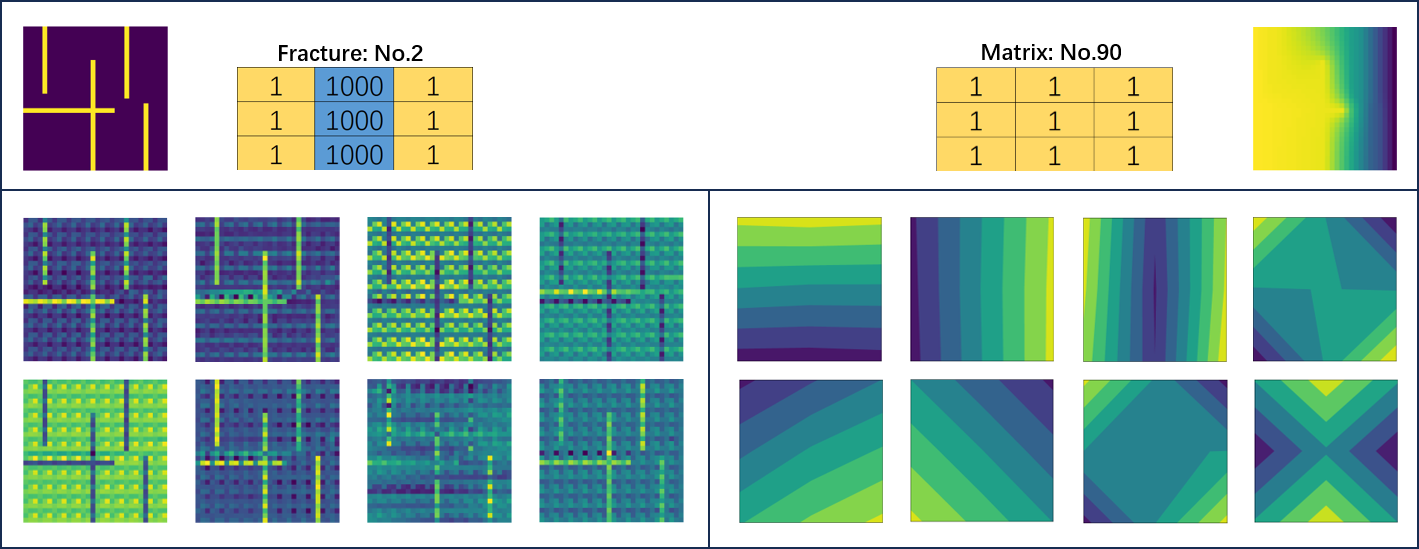}
	\caption{\textbf{Results of total operator sampling from validation set}. Top: permeability field, reconstructed pressure fields, and example coarse grid with and without fracture. Bottom left: examples of outputs of the preconditioner. For each basis function model we sampled the No.8 and No.16 one (there are totally 32 outputs). Bottom right: reconstructed multiscale basis functions. The up line refers to the fractured grid, down line refers to the matrix grid. }
	\label{fig:results_val}
\end{figure}

We visualized the reconstructed pressure fields using heatmaps. \hyperref[fig:results_comparison]{Fig.\ref{fig:results_comparison}} illustrates the reconstruction results for a specific sample using different combinations of basis functions. The left side of each image represents the numerical solution obtained via the mixed GMsFEM, while the right side shows the deep learning solutions corresponding to different combinations. From these visualizations, it can be observed that for pressure field data with distinct features, the proposed model is capable of accurately learning and reconstructing those features.

\begin{figure}[h]
	\centering
	\includegraphics[width=1.0\linewidth]{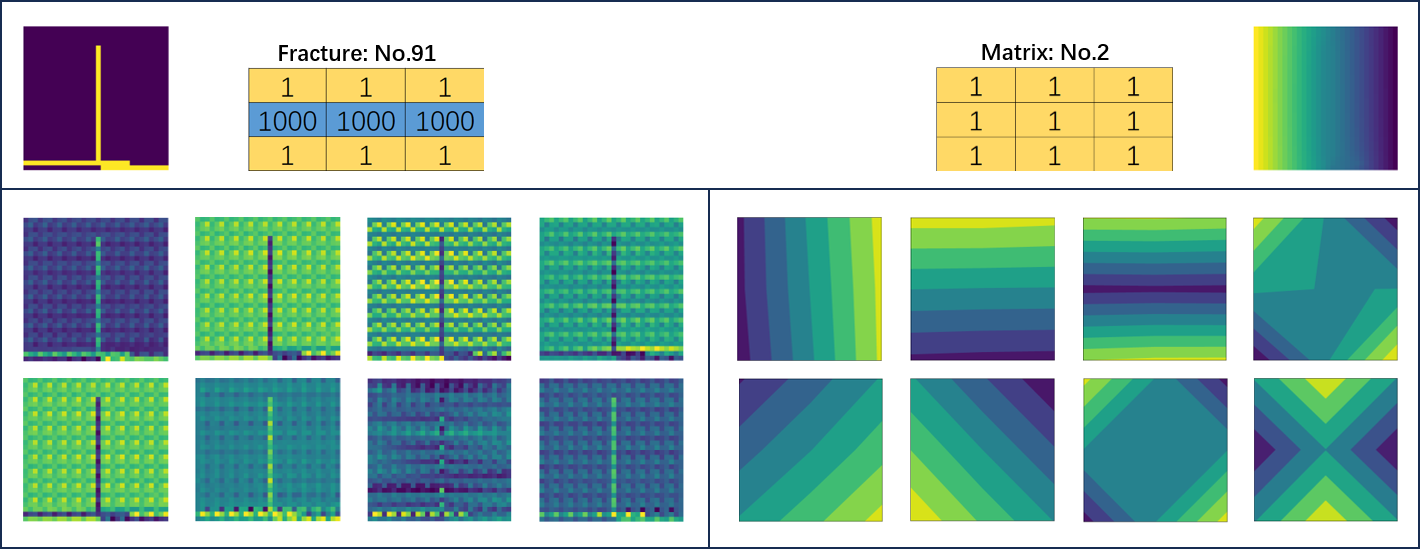}
	\caption{\textbf{Results of total operator sampling from testing set}. Top: permeability field, reconstructed pressure fields, and example coarse grid with and without fracture. Bottom left: examples of outputs of the preconditioner. Bottom right: reconstructed multiscale basis functions. The up line refers to the fractured grid, down line refers to the matrix grid. }
	\label{fig:results_test}
\end{figure}

To comprehensively demonstrate the reconstruction capabilities of our method and the output characteristics of several key modules, we combined and presented these components. The corresponding results can be observed in \hyperref[fig:results_val,fig:results_test,fig:results_train]{Fig.\ref{fig:results_val}, \ref{fig:results_test} and \ref{fig:results_train}}. These visualizations include the permeability fields of high-contrast fractured porous media, the pressure field solutions of the Darcy equation, partial outputs of the preconditioner, and the reconstructed multiscale basis functions on the corresponding grids. This part of the results can fully show some key outputs of our proposed model and reflect the effectiveness of our model to a certain extent.

\section{Discussion and Conclusion}
\label{sec:discussion}
In this paper, we proposed a hybrid-model-training two-stage method, developing an innovative neural operator to solve Darcy problem with high-contrast coefficients under the skeleton of mixed GMsFEM. In the first stage, traditional data-driven training method was hired to reconstruct the multiscale basis functions. For the purpose to eliminate the noise and expand the size of input features, we developed a domain transform-based preconditioner, using FNO to process the high-contrast permeability fields, together with a hierarchical multiscale feature extractor to capture the information contained. In the second stage, to efficiently solving the pressure field, we used physics-informed training method to achieve our purpose with a small-sample dataset. To utilize the reconstructed basis functions, we concatenate them with permeability fields as the model input. They will be processed by both local (symmetrical residual connections-based accelerator with a V-shaped CNN) and global (Transformer-based for capture spatial information and long-range dependencies) information learner. Both stages achieved good results in the reconstruction of multiscale basis functions and pressure fields.

Compared to existing research, our proposed model demonstrates superior performance. For the reconstruction of basis functions, since our model is primarily trained in a data-driven manner, the evaluation metrics focusing on the data itself are of particular importance. On the test set, our model achieved mean squared error (MSE) values of 0.0036, 0.0007, 0.0021, and 0.0027, and corresponding R2 values of 0.9716, 0.9148, 0.9366, and 0.9542 for different basis functions. In contrast, the results from \cite{choubineh2022innovative} reported MSE values of 0.0466, 0.0743, 0.0184, and 0.0154, with corresponding R$^2$ values of 0.8083, 0.6445, 0.8762, and 0.7625. These results indicate that our model achieves better data fitting and exhibits stronger generalization capabilities.

For the reconstruction of pressure fields in the second stage, our model achieved R$^2$ values ranging from 0.9032 to 0.9111 for different basis function combinations. In comparison, \cite{choubineh2023deep} reported R$^2$ values ranging from 0.8456 to 0.9191. While the R2 values are comparable, we place greater emphasis on whether the results align with real-world physical laws. To address this, we incorporated the average absolute percentage residual (AAPR) as an additional evaluation metric. In this context, all combinations exhibited AAPR values on the order of $1\times 10^{-4}$ or smaller. Despite the similar R2values, our model achieves a higher lower bound for R2, indicating that it delivers comparable or better reconstruction performance while better adhering to physical laws.

While the proposed framework performs well in reconstructing multiscale basis functions and pressure fields, several limitations should be noted. The reliance on large, high-quality datasets in the first stage may limit its applicability in scenarios where data is scarce or expensive to acquire. In addition, the inclusion of physical constraints in the second stage, while beneficial for maintaining physical consistency, adds complexity to the training process and requires careful tuning of hyperparameters. The computational overhead introduced by the Transformer module also presents challenges when scaling to larger datasets or higher dimensional domains.

Future research should focus on addressing these limitations by leveraging strategies such as data-efficient approaches, including generation models or transfer learning, to reduce data dependency and enhance generalization. A lightweight Transformer architecture can be explored to mitigate computational costs, while extending the framework to time-dependent or multiphase flow problems will greatly broaden its scope. In addition, combining uncertainty quantization technology can improve the robustness and reliability of the model, so as to achieve more practical applications in real-world scenarios.

\section*{CRediT Authors Contributions Statement}
\textbf{Peiqi Li}: Conceptualization, Methodology, Project administration, Visualization, Writing-original draft \& review \& editing. \textbf{Jie Chen}: Conceptualization, Data curation, Formal analysis, Methodology, Software, Supervision, Validation, Writing-review \& editing.

\section*{Data and Code Availability}
Due to proprietary considerations, both the data set and the analysis code used in this study are not publicly available. Researchers interested in accessing the data or the code may contact the corresponding author for more information.

\section*{Declaration of Competing Interests}
The authors confirm that they have no conflict of interest in relation to the content of this study. 

\newpage
\appendix
\section{Knowledge of Mixed Generalized Multiscale Finite Element Methods for Darcy Equation}
\label{appendix:mixed gmsfem}

We consider the following first-order Darcy's Flow problem with high-contrast coefficients and nonhomogeneous boundary condition in Lipschitz continuous domain:
\begin{align}
	\begin{cases}
		\kappa^{-1}\mathbf{u} + \nabla p &= 0, \quad \text{in} \, \Omega \\
		\nabla \cdot \mathbf{u} &= f, \quad \text{in} \, \Omega \\
		\mathbf{u} \cdot \mathbf{n} &= g, \quad \text{on} \, \partial \Omega 
	\end{cases}
\end{align}
where $\kappa$ is high-contrast heterogeneous permeability field, $\mathbf{u}$ and $p$ are the Darcy velocity and pressure, respectively. $f$ is the known source term, $g$ is given normal component of Darcy velocity on the boundary, $\Omega=[0,1]^2$ is the computational domain, and $\mathbf{n}$ is the outward unit norm vector on the boundary.

Let $\mathcal{T}^H$ denote a conforming partition of $\Omega$ into finite elements with coarse-grid size $H$ and $\mathcal{T}^h$ denote the fine-grid partition of $\Omega$ into non-overlapping elements with size $h$. Define $\mathcal{E}^H:=\bigcup_{i=1}^{N_e} E_i$ and $\mathcal{E}^h:=\bigcup_{i=1}^{M_e}e_i$ as the set of all edges in the coarse ad fine mesh $\mathcal{T}^H$ and $\mathcal{T}^h$, where $N_e$ and $M_e$ refer to the number of coarse and fine edges, respectively. 

Define
\begin{equation}
	L^2(\Omega)=\{ v: \text{$v$ is defined on $\Omega$~\text{and}~$\int_{\Omega} v^2 \text{d}x < \infty$ \}} \notag
\end{equation}

Use the space $\text{H}(\text{div},\Omega)=\{ \mathbf{v}=(\mathbf{v}_1,\mathbf{v}_2)\in (L^2(\Omega))^2:\ \nabla \cdot \mathbf{v} \in L^2(\Omega) \}$. Define
\begin{equation}
	\text{V}=\text{H}(\text{div},\Omega), \quad \text{W}=L^2(\Omega) \notag
\end{equation}

Define the mixed finite element spaces
\begin{align}
	\text{V}_h &= \{ \mathbf{v}_h \in \text{V}:\ \mathbf{v}_h|_t=(b_tx_1+a_t, d_tx_2+c_t),\ a_t, b_t, c_t, d_t \in \mathbb{R}, t \in \mathcal{T}^g \} \notag \\
	\text{W}_h &= \{ \text{w}_h \in \text{W}: \text{$\text{w}_h$ is a constant on each element in $\mathcal{T}^h$} \} \notag
\end{align}

The normal components of $\mathbf{v}_h$ are continuous across the interior edges in $\mathcal{T}^h$. Then the solution $(\mathbf{u}_h,p_h) \in (\text{V}_h,\text{W}_h)$ on the fine grid will satisfy
\begin{align}
	\int_{\Omega}\kappa^{-1} \mathbf{u}_h \cdot \mathbf{v}_h - \int_{\Omega} \text{div}(\mathbf{v}_h)p_h&=0, \quad \quad \quad \ \  \forall \mathbf{v}_h \in \text{V}_h^0, \\
	\int_{\Omega} \text{div}(\mathbf{u}_h)\text{w}_h&=\int_{\Omega}f\text{w}_h, \quad \forall \text{w}_h \in \text{W}_h \notag
\end{align}
where $\text{V}_h^0=\{ \mathbf{v}_h \in \text{V}_h: \mathbf{v}_h \cdot n = 0\ \text{on $\partial \Omega$} \}$. Then the system above can be written as the matrix form
\begin{align}
	\text{M}_{\text{fine}}\text{U}_h+\text{B}_{\text{fine}}\text{P}_h &= 0 \\
	\text{B}_{\text{fine}}^T\text{U}_h &= \text{F}_h \notag
\end{align}

The approximation of pressure field will then be found in the multiscale finite element space between the coarse-grid and fine-grid space. The multiscale space $\text{W}^{\text{snap}}$ for $\text{p}$ is defined as the linear span of all local basis functions:
\begin{equation}
	\text{W}_H=\bigoplus_{\mathcal{T}^H}\{ \Psi_i\} \notag
\end{equation}

There are three ways to construct the snapshot space:
\begin{enumerate}
	\item \textbf{Pressure Snapshot Space on the Find Grid}:
	\begin{equation} \text{W}^{\text{snap}}=\{\phi_i^{\text{snap}}: \text{piecewise constant on the fine grid} \} \notag
	\end{equation}
	\item \textbf{Local Dirichlet Problem}: Solve the local problem for each coarse grid unit $T_i$:
	\begin{align}
		\begin{cases}
			\kappa^{-1}\mathbf{u}_j^{(i)}+\nabla p_j^{(i)} = 0, \quad \text{in} \ T_i, \\
			\text{div}(\mathbf{u}_j^{(i)}) = 0, \quad \quad \quad\ \ \  \text{in}\ T_i, \\
			p_j^{(i)}=\delta_j^{(i)}= \begin{cases}
				1\  \text{in}\ e_j, \\
				0\ \text{on other fine-grid edges on}\ \partial \Omega
			\end{cases},j=1,2,\cdots J_i.
		\end{cases}
	\end{align}
	this problem together with Dirichlet boundary condition can be solved numerically on the find grid $T_i$ by lowest-order Raviart-Thomas element s.t. resulting $\text{p}_j^{(i)} \in \text{W}_h$.
	
	\item \textbf{Local Neumann Problem}: solve the local problem for each coarse grid unit $T_i$:
	\begin{align}
		\begin{cases}
			\kappa^{-1}\mathbf{u}_j^{(i)}+\nabla p_j^{{(i)}} &=0,\ \ \text{in} \ T_i, \\
			\nabla \cdot \mathbf{u}_j^{(i)}&=\alpha_j, \  \text{in}\ T_i, \\
			\frac{\partial p_j^{(i)}}{\partial \mathbf{n}_i}&=\delta_j, \text{on}\  \partial \Omega
		\end{cases}
	\end{align}
	
	where $\mathbf{n}_i$ refers to an outward unit norm vector on $\partial T_i$, $\alpha_j^{(i)}$ is chosen s.t. the compatibility condition $\int_{T_i}\alpha_j^{(i)}=\int_{\partial T_i}\delta_j^{(i)}$ is satisfied. 
	
	\textit{Remark}: In our case, $p$ is unique up to an additive constant.
\end{enumerate}

Snapshot space will then be constructed by the solutions of above local problems:
\begin{equation}
	\text{W}^{\text{snap}}=\text{span}\{\phi_j^{\text{snap}}:j=1,2,\cdots,J_i,\ \forall T_i \in T_H \} \notag
\end{equation}
where $J_i$ is the number of fine element edges in the coarse element boundary.

For each coarse-grid element, in the snapshot $\text{W}^{\text{snap}}$, reduce the spatial dimension by a local spectral problem:
\begin{equation}
	a_i(p,\text{w})=\lambda s_i(p,\text{w}),\ \forall \text{w}\in \text{W}^{\text{snap}}
\end{equation}
where $a_i(p,\text{w})=\sum_{e\in\mathcal{E}_h^0}\kappa_e[p_h]_e[\text{w}_h]_e$ is the stiffness matrix representation of snapshot space, $s_i(p,\text{w})=\int_{T_i}kp\text{w}$ is the mass matrix representation. The discretization form of spectral problem can be written as:
\begin{equation}
	\text{A}_i\text{Z}_k=\lambda_k\text{S}_i\text{Z}_k
\end{equation}
where $\text{A}_i$ and $\text{S}_i$ are the stiffness matrix and mass matrix, respectively. $\lambda_k$ refers to the eigenvalue and $\text{Z}_k$ is the corresponding eigenvector.

The eigenvalues are arranged in ascending order, and the first $l_i$ eigenvalues corresponding to the smallest eigenvalues are selected to generate the offline basis functions:
\begin{equation}
	\phi_k^{\text{off}}=\sum_{j=1}^{J_i}\text{Z}_{k,j}\phi_j^{\text{snap}}
\end{equation}
where $\text{Z}_{k,j}$ represents the components of the eigenvector $\text{Z}_k$.

The offline basis functions of all relevant elements are combined to construct the global offline space:
\begin{equation}
	\text{W}^{\text{off}}=\text{span}\{\phi_k^{\text{off}}:k=1,2,\cdots,l_i,\ \forall T_i \in T_H \} = \text{span}\{\phi_m^{\text{off}}:m=1,2,\cdots,M^{\text{off}}\} \notag
\end{equation}
where $M^{\text{off}}=\sum_{T_i\in T_H}l_i$.

The mixed system in the offline space is given as:
\begin{align}
	\int_{\Omega}\kappa^{-1}\mathbf{u}_\text{H}\cdot \mathbf{v}_{\text{H}}-\int_{\Omega}(\nabla \cdot \mathbf{v}_{{\text{H}}})p_{\text{H}}&=0,\ \forall \mathbf{v}_{\text{H}}\in \text{V}_h, \\
	\int_{\Omega} (\nabla \cdot \mathbf{u}_{\text{H}})\text{w}_{\text{H}}&=\int_{\Omega}f\text{w}_{\text{H}},\ \forall \text{w}_{\text{H}}\in \text{W}^{{\text{off}}} \notag
\end{align}

After solving the coarse-grid pressure field $p_{\text{H}}$, the fine-grid pressure field $p_{\text{h}}$ will be obtained by interpolation using the offline basis functions:
\begin{equation}
	p_h=\text{R}_{\text{off}}\text{P}_{\text{H}}
\end{equation}
where $\text{R}_{\text{off}}$ is the mapping from the offline space to the fine-grid space.

\newpage
\section{Examples of Our Computation Results}
\label{appedix:examples}

\begin{figure}[h]
	\centering
	\includegraphics[width=0.9\linewidth]{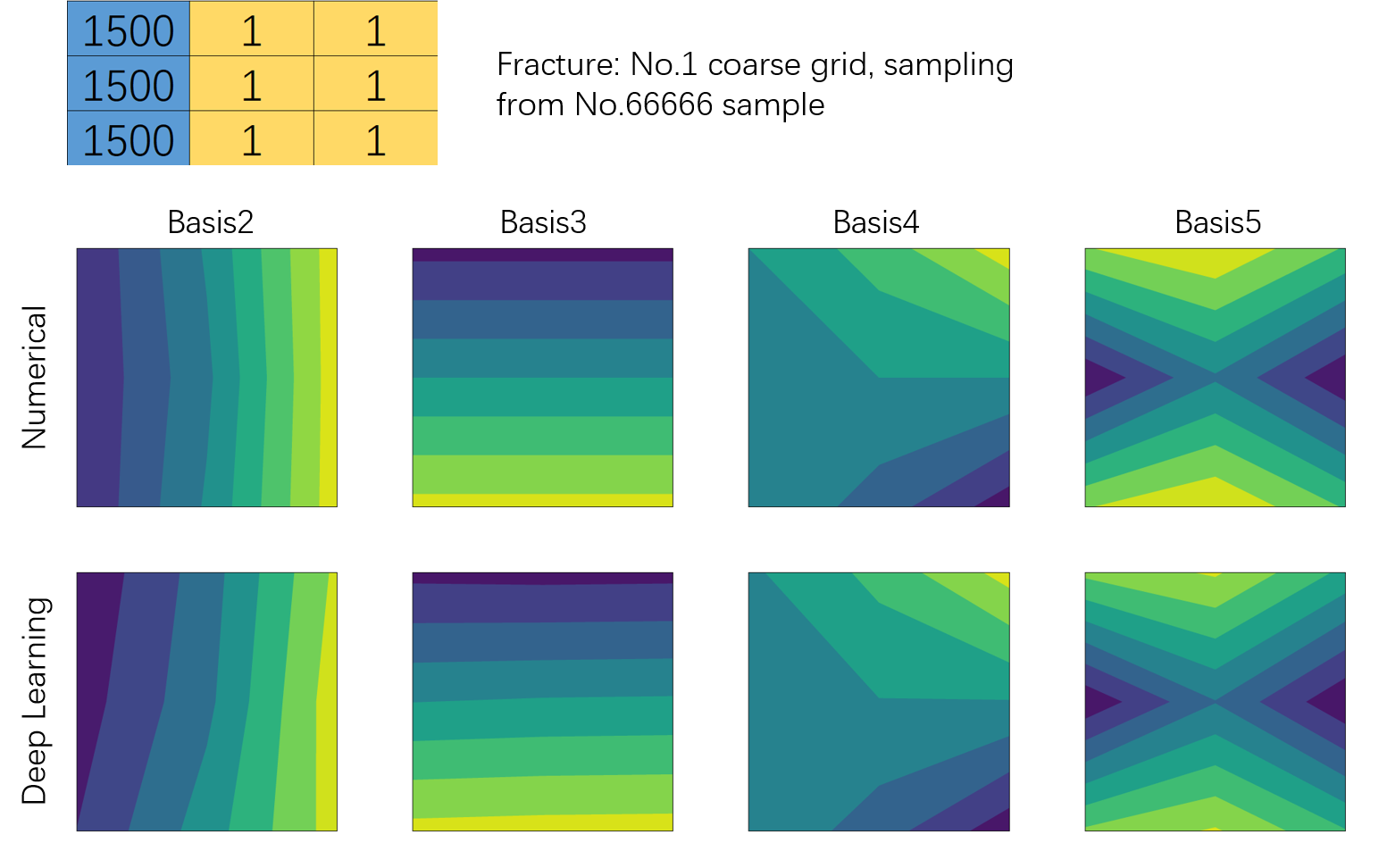}
	\caption{Example of Reconstruction for basis functions, sampling from No.1 coarse grid of No.66666 training sample (with fractures). Top: permeability value for each element, (1-5): matrix, (500-2000):fracture. Bottom: comparison between actual and reconstructed basis functions.}
	\label{fig:train fracture}
\end{figure}

\newpage

\begin{figure}[h]
	\centering
	\includegraphics[width=1.0\linewidth]{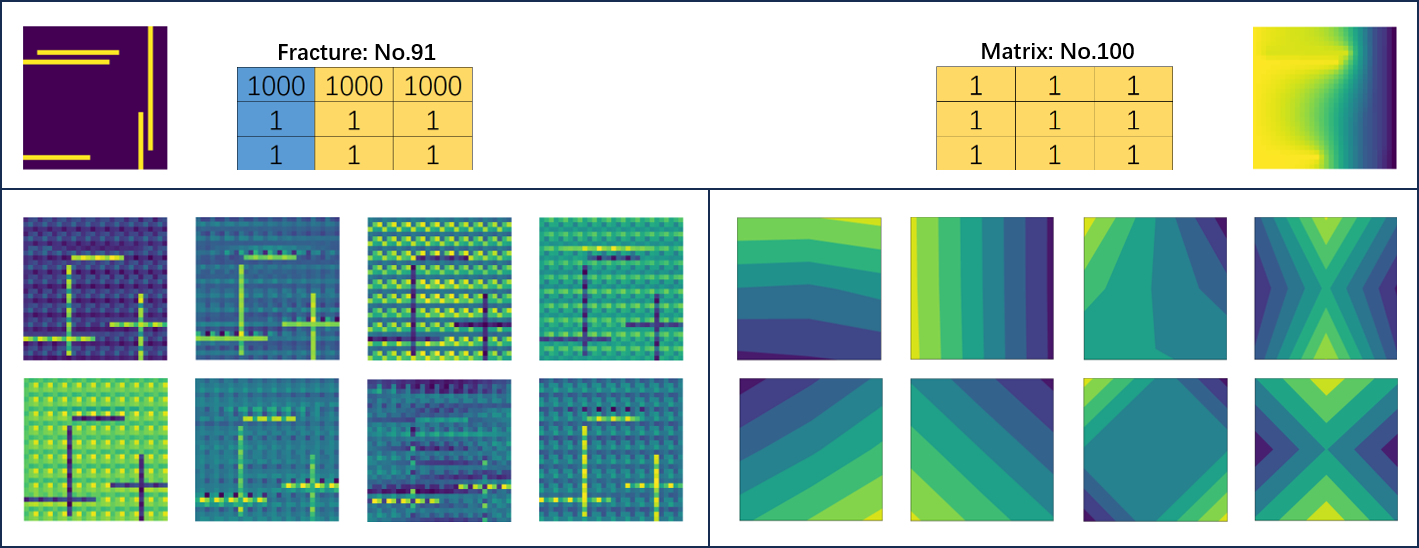}
	\caption{\textbf{Results of total operator sampling from training set}. Top: permeability field, reconstructed pressure fields, and example coarse grid with and without fracture. Bottom left: examples of outputs of the preconditioner. For each basis function model we sampled the No.8 and No.16 one (there are totally 32 outputs). Bottom right: reconstructed multiscale basis functions. The up line refers to the fractured grid, dowm line refers to the matrix grid. }
	\label{fig:results_train}
\end{figure}

\newpage
\begin{figure}[h]
	\centering
	\includegraphics[width=1.0\linewidth]{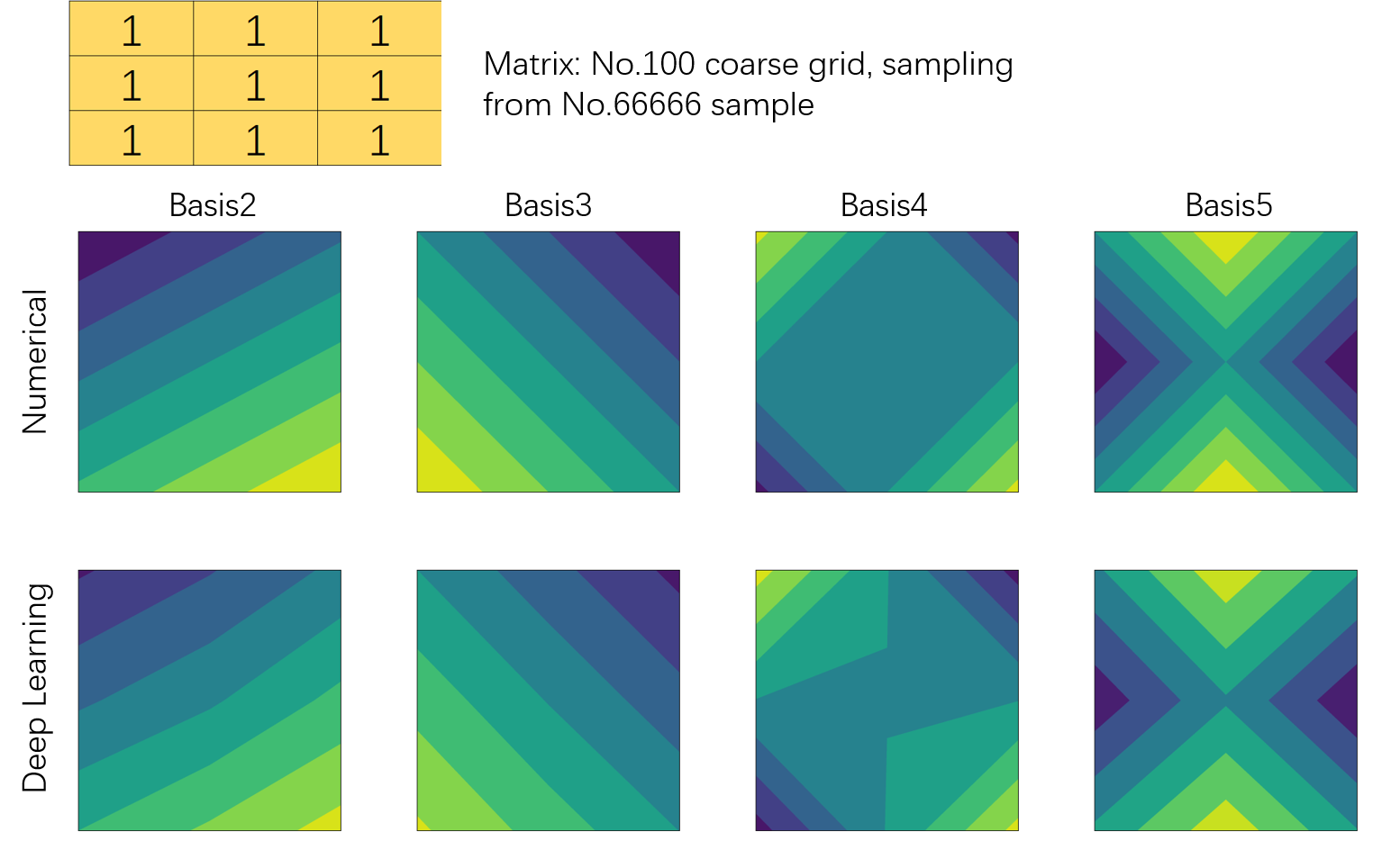}
	\caption{Example of Reconstruction for basis functions, sampling from No.100 coarse grid of No.66666 training sample (without fractures).}
	\label{fig:train matrix}
\end{figure}

\newpage
\bibliographystyle{elsarticle-num} 
\bibliography{main.bib}

\begin{thebibliography}{10}
\expandafter\ifx\csname url\endcsname\relax
  \def\url#1{\texttt{#1}}\fi
\expandafter\ifx\csname urlprefix\endcsname\relax\def\urlprefix{URL }\fi
\expandafter\ifx\csname href\endcsname\relax
  \def\href#1#2{#2} \def\path#1{#1}\fi

\bibitem{kippe2008comparison}
V.~Kippe, J.~E. Aarnes, K.-A. Lie, A comparison of multiscale methods for
  elliptic problems in porous media flow, Computational Geosciences 12~(3)
  (2008) 377--398.

\bibitem{efendiev2007multiscale}
Y.~Efendiev, T.~Hou, Multiscale finite element methods for porous media flows
  and their applications, Applied Numerical Mathematics 57~(5-7) (2007)
  577--596.

\bibitem{aarnes2006adaptive}
J.~E. Aarnes, Y.~Efendiev, An adaptive multiscale method for simulation of
  fluid flow in heterogeneous porous media, Multiscale Modeling \& Simulation
  5~(3) (2006) 918--939.

\bibitem{dehkordi2013multi}
M.~M. Dehkordi, M.~T. Manzari, A multi-resolution multiscale finite volume
  method for simulation of fluid flows in heterogeneous porous media, Journal
  of Computational Physics 248 (2013) 339--362.

\bibitem{jenny2003multi}
P.~Jenny, S.~Lee, H.~A. Tchelepi, Multi-scale finite-volume method for elliptic
  problems in subsurface flow simulation, Journal of computational physics
  187~(1) (2003) 47--67.

\bibitem{hajibeygi2008iterative}
H.~Hajibeygi, G.~Bonfigli, M.~A. Hesse, P.~Jenny, Iterative multiscale
  finite-volume method, Journal of Computational Physics 227~(19) (2008)
  8604--8621.

\bibitem{hou1997multiscale}
T.~Y. Hou, X.-H. Wu, A multiscale finite element method for elliptic problems
  in composite materials and porous media, Journal of computational physics
  134~(1) (1997) 169--189.

\bibitem{efendiev2009multiscale}
Y.~Efendiev, T.~Hou, Multiscale finite element methods. surveys and tutorials
  in the applied mathematical sciences, Springer, New York 4 (2009) 1.

\bibitem{arbogast2007multiscale}
T.~Arbogast, G.~Pencheva, M.~F. Wheeler, I.~Yotov, A multiscale mortar mixed
  finite element method, Multiscale Modeling \& Simulation 6~(1) (2007)
  319--346.

\bibitem{arbogast2013multiscale}
T.~Arbogast, Z.~Tao, H.~Xiao, Multiscale mortar mixed methods for heterogeneous
  elliptic problems, Contemp. Math 586 (2013) 9--21.

\bibitem{chen2020generalized}
J.~Chen, E.~T. Chung, Z.~He, S.~Sun, Generalized multiscale approximation of
  mixed finite elements with velocity elimination for subsurface flow, Journal
  of Computational Physics 404 (2020) 109133.

\bibitem{kovachki2023neural}
N.~Kovachki, Z.~Li, B.~Liu, K.~Azizzadenesheli, K.~Bhattacharya, A.~Stuart,
  A.~Anandkumar, Neural operator: Learning maps between function spaces with
  applications to pdes, Journal of Machine Learning Research 24~(89) (2023)
  1--97.

\bibitem{raonic2024convolutional}
B.~Raonic, R.~Molinaro, T.~De~Ryck, T.~Rohner, F.~Bartolucci, R.~Alaifari,
  S.~Mishra, E.~de~B{\'e}zenac, Convolutional neural operators for robust and
  accurate learning of pdes, Advances in Neural Information Processing Systems
  36 (2024).

\bibitem{choubineh2022innovative}
A.~Choubineh, J.~Chen, F.~Coenen, F.~Ma, An innovative application of deep
  learning in multiscale modeling of subsurface fluid flow: Reconstructing the
  basis functions of the mixed gmsfem, Journal of Petroleum Science and
  Engineering 216 (2022) 110751.

\bibitem{choubineh2023deep}
A.~Choubineh, J.~Chen, D.~A. Wood, F.~Coenen, F.~Ma, Deep ensemble learning for
  high-dimensional subsurface fluid flow modeling, Engineering Applications of
  Artificial Intelligence 126 (2023) 106968.

\bibitem{li2020fourier}
Z.~Li, N.~Kovachki, K.~Azizzadenesheli, B.~Liu, K.~Bhattacharya, A.~Stuart,
  A.~Anandkumar, Fourier neural operator for parametric partial differential
  equations, arXiv preprint arXiv:2010.08895 (2020).

\bibitem{li2023fourier}
Z.~Li, D.~Z. Huang, B.~Liu, A.~Anandkumar, Fourier neural operator with learned
  deformations for pdes on general geometries, Journal of Machine Learning
  Research 24~(388) (2023) 1--26.

\bibitem{wang2024recent}
H.~Wang, Y.~Cao, Z.~Huang, Y.~Liu, P.~Hu, X.~Luo, Z.~Song, W.~Zhao, J.~Liu,
  J.~Sun, et~al., Recent advances on machine learning for computational fluid
  dynamics: A survey, arXiv preprint arXiv:2408.12171 (2024).

\bibitem{vaswani2017attention}
A.~Vaswani, Attention is all you need, Advances in Neural Information
  Processing Systems (2017).

\bibitem{meng2023transformer}
Y.~Meng, J.~Jiang, J.~Wu, D.~Wang, Transformer-based deep learning models for
  predicting permeability of porous media, Advances in Water Resources 179
  (2023) 104520.

\bibitem{li2024transformer}
Z.~Li, T.~Liu, W.~Peng, Z.~Yuan, J.~Wang, A transformer-based neural operator
  for large-eddy simulation of turbulence, arXiv preprint arXiv:2403.16026
  (2024).

\bibitem{zhao2023pinnsformer}
Z.~Zhao, X.~Ding, B.~A. Prakash, Pinnsformer: A transformer-based framework for
  physics-informed neural networks, arXiv preprint arXiv:2307.11833 (2023).

\bibitem{li2024physics}
Z.~Li, H.~Zheng, N.~Kovachki, D.~Jin, H.~Chen, B.~Liu, K.~Azizzadenesheli,
  A.~Anandkumar, Physics-informed neural operator for learning partial
  differential equations, ACM/JMS Journal of Data Science 1~(3) (2024) 1--27.

\bibitem{pang2019fpinns}
G.~Pang, L.~Lu, G.~E. Karniadakis, fpinns: Fractional physics-informed neural
  networks, SIAM Journal on Scientific Computing 41~(4) (2019) A2603--A2626.

\bibitem{jagtap2020adaptive}
A.~D. Jagtap, K.~Kawaguchi, G.~E. Karniadakis, Adaptive activation functions
  accelerate convergence in deep and physics-informed neural networks, Journal
  of Computational Physics 404 (2020) 109136.

\bibitem{raissi2019physics}
M.~Raissi, P.~Perdikaris, G.~E. Karniadakis, Physics-informed neural networks:
  A deep learning framework for solving forward and inverse problems involving
  nonlinear partial differential equations, Journal of Computational physics
  378 (2019) 686--707.

\bibitem{fukunaga1970application}
K.~Fukunaga, W.~L. Koontz, Application of the karhunen-loeve expansion to
  feature selection and ordering, IEEE Transactions on computers 100~(4) (1970)
  311--318.

\end{thebibliography}
\end{document}